\documentclass[10pt,twocolumn,letterpaper]{article}
\pdfoutput=1

\usepackage{cvpr}              % To produce the CAMERA-READY version
\usepackage[dvipsnames]{xcolor}

\definecolor{cvprblue}{rgb}{0.21,0.49,0.74}
\definecolor{i3d}{HTML}{749f83}
\definecolor{timesformerk400}{HTML}{d48265}
\definecolor{videomaek400}{HTML}{67acc7}
\definecolor{videomaek710}{HTML}{61a0a8}
\newcommand{\myparagraph}[1]
{
\vspace{3mm}\noindent\textbf{#1}
}

\usepackage[pagebackref,breaklinks,colorlinks,citecolor=cvprblue]{hyperref}

\usepackage{graphicx}
\usepackage{amsmath}
\usepackage{amssymb}
\usepackage{booktabs}
\usepackage{rotating}

\usepackage{color}
\usepackage{makecell}

\definecolor{tabhigh}{rgb}{0.9, 0.25, 0.25}
\definecolor{tabmid}{rgb}{0.9, 0.6, 0.05}
\definecolor{tablow}{rgb}{0.4, 0.8, 0.35}

\usepackage{tikz}
\usepackage{pgfplots}
\pgfplotsset{compat=1.18}
\usepackage{graphicx}
\usepackage{amsmath}
\usepackage{amssymb}
\usepackage{booktabs}
\usepackage{url}
\usepackage{epsfig}
\usepackage{lipsum}
\usepackage{amsthm}
\usepackage{comment}
\usepackage{multirow}
\usepackage{subcaption}
\usepackage{microtype}
\usepackage{xspace}
\usepackage{enumitem}
\usepackage{graphbox}
\usepackage{epigraph}
\usepackage{wrapfig}
\usepackage{textcomp}  % Required for encoding \textbigcircle
\usepackage{scalerel}  % Required for emoji \scalerel
\usepackage{animate}
\usepackage{colortbl}
\usepackage{tabularx,booktabs}
\usepackage{pifont}
\def\paperID{4702} % *** Enter the Paper ID here
\def\confName{CVPR}
\def\confYear{2024}

\title{
On the Content Bias in Fréchet Video Distance
}

\author{
Songwei Ge\textsuperscript{1}
\qquad
Aniruddha Mahapatra\textsuperscript{2,3}
\qquad
Gaurav Parmar\textsuperscript{2}
\\
Jun-Yan Zhu\textsuperscript{2}
\qquad
Jia-Bin Huang\textsuperscript{1}\\
\vspace{-3.5mm}\\
\textsuperscript{1}University of Maryland, College Park\qquad
\textsuperscript{2}Carnegie Mellon University\qquad
\textsuperscript{3}Adobe Research\\
\url{https://content-debiased-fvd.github.io/}
\vspace{2mm}
}

\begin{document}
\twocolumn[{%
\maketitle
\renewcommand\twocolumn[1][]{#1}%
\setlength{\tabcolsep}{0.5pt}
\renewcommand{\arraystretch}{0.5}

\noindent % Ensures the table starts at the left edge of the text area
\begin{tabular}{lc}
\begin{minipage}[c][\height][c]{0.03\textwidth}
    \centering
    \begin{turn}{90}
        Videos
    \end{turn}
\end{minipage} &
\begin{minipage}[c][\height][c]{0.97\textwidth}
    % \begin{tabular}{c@{\vrule width 1pt}cc}
    \begin{tabular}{c@{\hspace{5pt}}!{\vrule width 0.8pt}@{\hspace{5pt}}cc}
    \animategraphics[autoplay, loop, width=0.3207\textwidth]{16}{videos/teaser/clean/}{0001}{0016} & 
    \animategraphics[autoplay, loop, width=0.3207\textwidth]{16}{videos/teaser/spatial/}{0001}{0016} & 
    \animategraphics[autoplay, loop, width=0.3207\textwidth]{16}{videos/teaser/temporal/}{0001}{0016} 
    \end{tabular}
\end{minipage}
\end{tabular}

\vspace{25pt}

\noindent % Ensures the table starts at the left edge of the text area
\begin{tabular}{lc}
\begin{minipage}[c][\height][c]{0.03\textwidth}
    \centering
    \begin{turn}{90}
        x-t slice
    \end{turn}
\end{minipage} &
\begin{minipage}[c][\height][c]{0.97\textwidth}
\includegraphics[width=0.99\linewidth]{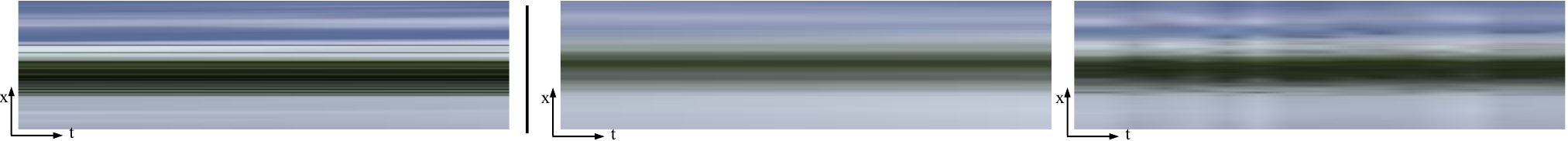}
\end{minipage}
\end{tabular}

\vspace{5pt} % Adds a bit of vertical space between the tables

\noindent % Ensures the table starts at the left edge of the text area
\renewcommand{\arraystretch}{1.0}
\begin{tabularx}{\textwidth}{>{\centering\arraybackslash}m{0.02\textwidth}@{\hspace{5pt}}>{\centering\arraybackslash}m{0.3\textwidth}@{\hspace{5pt}}>{\centering\arraybackslash}m{0.33\textwidth}@{\hspace{4pt}}>{\centering\arraybackslash}m{0.31\textwidth}} 
    & \small Reference Videos & 
    \small (a) Medium Spatial \&  No Temporal Corruption &
    \small (b) Small Spatial \& Severe Temporal Corruption \\
    & & \small FVD=317.10 & \small FVD=310.52
\end{tabularx}
\renewcommand{\arraystretch}{0.5}
\vspace{-2mm}
\captionof{figure}{\textbf{FVD is biased towards per-frame quality than temporal consistency.} 
FVD~\cite{unterthiner2019fvd}, a commonly used video generation evaluation metric, should ideally capture \textit{both} spatial and temporal aspects. However, our experiments reveal a strong bias toward individual frame quality. 
(a) First, we apply mild spatial distortions through local warping, which results in an FVD score of 317.10.
(b) Next, we induce slightly less spatial corruptions but \textit{severe} temporal inconsistencies by altering each frame differently.
These changes create artifacts that are noticeable to humans and evident in the spatiotemporal x-t slice, as seen in the bottom row, but surprisingly lead to a lower (better) FVD score of 310.52. 
This discrepancy highlights the metric's bias towards individual frame quality. 
\emph{We encourage readers to view the videos with Acrobat Reader or visit our website to observe the inconsistencies.}}
\vspace{3mm}
\label{fig:teaser}

}]
\begin{abstract}
\vspace{-2mm}
Fréchet Video Distance (FVD), a prominent metric for evaluating video generation models, is known to conflict with human perception occasionally.  In this paper, we aim to explore the extent of FVD's bias toward per-frame quality over temporal realism and identify its sources. We first quantify the FVD's sensitivity to the temporal axis by decoupling the frame and motion quality and find that the FVD increases only slightly with large temporal corruption.  We then analyze the generated videos and show that via careful sampling from a large set of generated videos that do not contain motions, one can drastically decrease FVD without improving the temporal quality.  Both studies suggest FVD's bias towards the quality of individual frames.  We further observe that the bias can be attributed to the features extracted from a supervised video classifier trained on the content-biased dataset.  We show that FVD with features extracted from the recent large-scale self-supervised video models is less biased toward image quality.  Finally, we revisit a few real-world examples to validate our hypothesis.
\vspace{-2mm}
\end{abstract}

\section{Introduction}
\label{sec:intro}

Video generation~\cite{singer2022make,ho2022imagen,ge2023preserve,blattmann2023align,girdhar2023emu,bartal2024lumiere,videoworldsimulators2024} has recently accomplished unprecedented advances driven by the scalable models~\cite{ho2020denoising,song2021denoising} and growing training data~\cite{schuhmann2022laionb,bain2021frozen}. 
With rapid progress, it is increasingly crucial to evaluate the model performance accurately. 
Despite the fruitful literature on assessing video quality~\cite{sinno2018large,wang2004video,seshadrinathan2010study,li2019quality} and designing image generation evaluation metrics~\cite{heusel2017gans,sajjadi2018assessing,kynkaanniemi2019improved,naeem2020reliable,parmar2022aliased}, automatically evaluating the quality and diversity of the generated videos, has received less attention~\cite{liu2023evalcrafter,huang2023vbench}. 
In this paper, we focus on analyzing the bias of Fréchet Video Distance (FVD)~\cite{unterthiner2019fvd},  one of the most frequently used metrics for video evaluation. 

FVD extends the image generation metric Fréchet Inception Distance (FID)~\cite{heusel2017gans} to measure the quality and diversity of generated videos with respect to the training set.  
Given $N$ features $\mathbf{f}_i$  for a set of generated and real videos, which are column vectors extracted from a pretrained video network, we fit a multivariate Gaussian with the mean $\mathbf{\mu}=\frac{1}{N} \sum_i \mathbf{f}_i$, and covariance  $\mathbf{\Sigma} = \frac{1}{N} \sum_i(\mathbf{f}_i-\mathbf{\mu})(\mathbf{f}_i-\mathbf{\mu})^T$. 
The performance of the video generator is then measured as the Fréchet distance~\cite{dowson1982frechet} between the two Gaussian distributions:
\begin{equation}
    \label{eq:fvd}
    \text{FVD} = \|\mathbf{\mu}_r-\mathbf{\mu}_g\|^2_2 + \text{Tr}\left(\mathbf{\Sigma}_r + \mathbf{\Sigma}_g - 2 \left(\mathbf{\Sigma}_r\mathbf{\Sigma}_g\right)^{\frac{1}{2}}\right),
\end{equation}
where $(\mathbf{\mu}_r,\mathbf{\Sigma}_r)$ and $(\mathbf{\mu}_g,\mathbf{\Sigma}_g)$ denote the mean and covariance for real and generated data. 

Earlier studies have confirmed that FVD reliably reflects the model performance in various cases, such as training convergence~\cite{yu2023video}, hyperparameter tuning~\cite{ho2022imagen,brooks2022generating}, and architecture design~\cite{ho2022video,skorokhodov2021stylegan}. 
However, several recent studies have reported cases where FVD scores contradict human judgment~\cite{skorokhodov2021stylegan,ge2022long,brooks2022generating}. 
A recurrent argument is that FVD tends to value the image quality of individual frames more than the realism of motion.  
We refer to such bias as the \emph{content bias}, which is inspired by the video generation works in decoupling content and motion~\cite{Tulyakov_2018_CVPR,tian2021a,villegas2017decomposing,pmlr-v80-jang18a,Wang_2020_CVPR}.

As shown in Figure~\ref{fig:teaser}, we motivate our analysis with a simple, controlled setting,  where the metric diverges from human perception when weighing spatial and temporal qualities. 
Specifically, given a set of reference videos, we create two sets of distortion. 
In (a), we locally warp the frames in each video uniformly,  while in (b), we distort the frames differently but with slightly reduced severity. 
The latter creates additional temporal artifacts. 
The FVD metric, however, favors video set (b), while most humans would pick video set (a) to be more similar to the reference videos due to the significant temporal inconsistency presented in video set (b).

Building upon this simple example, we present the first systematic study to quantify the content bias and understand its impact using both synthetic and real-world settings.
We first distort videos so that the frame quality deteriorates to the same level 
while the temporal consistency is either intact or, in the other case, significantly decreased. 
By comparing FVDs on these distorted videos, we can quantify the relative sensitivity of the FVD metric to the temporal consistency. 
Next,  following the previous work on FID analysis~\cite{kynkaanniemi2022role}, we probe the perceptual null space in the FVD metric. 
Without improving the temporal quality of the generated videos, we can still greatly reduce the FVD scores. Lastly, we revisit a few real-world examples where FVD presents a notable content bias.

Where does the content bias originate from?
Previous studies show that the alignment of the FID metric to human perception depends on the choice of the extracted features~\cite{morozov2020self,borji2022pros,alfarra2022robustness,kynkaanniemi2022role}.
In practice, FVD employs an Inflated 3D ConvNet (I3D) model~\cite{carreira2017quo}, originally trained for action recognition on the Kinetics-400 dataset~\cite{carreira2017quo}. 
The feature space is formed as the output of the logit layer.  
As a result, these features focus on extracting the semantic information about human actions in the videos.

Using I3D features thus raises several practical concerns that could undermine the metric's reliability. 
First, the Kinetics dataset~\cite{carreira2017quo} predominantly comprises videos with humans as the protagonists. However, video content diverging from typical Kinetics-400 categories, 
such as time-lapse landscape videos~\cite{xiong2018learning} or first-person riding and biking videos~\cite{brooks2022generating}, may not produce a meaningful feature representation. 
Second, the models trained on the Kinetics dataset are shown biased to the appearance of objects and backgrounds instead of motions~\cite{li2018resound,huang2018makes,wang2016temporal,choi2019can,liu2021no,Sevilla-Lara_2021_WACV}. 
For example, to recognize the action ``playing saxophone'', it is sufficient to detect the presence of a saxophone since this is the only category where a saxophone is presented. 
Therefore, the features may not capture the musician's motion. 
Previous works also show that descent classification accuracy can be achieved without modeling the temporal aspect~\cite{carreira2017quo,gberta_2021_ICML}.

To verify our hypothesis, we compute FVD scores using features extracted from a self-supervised model~\cite{wang2023videomaev2} trained on
diverse dataset and perform the same analysis. 
Overall, our experiments show that the FVD, computed with I3D features, is strongly biased to the content over the motion, while using features computed with a model trained in a self-supervised manner helps mitigate such bias to a large extent. 
Our evaluation code and data are available on \url{https://content-debiased-fvd.github.io/}.

\section{Related Work}
\label{sec:formatting}

\paragraph{Video generation.} 
Various types of generative models have been proposed for video generation such as GANs~\cite{vondrick2016generating,saito2017temporal,Tulyakov_2018_CVPR,tian2021a,Shen_2023_CVPR,villegas2017decomposing,Wang_2020_CVPR}, Autoregressive models~\cite{ranzato2014video,Yu_2023_CVPR,srivastava2015unsupervised,yan2021videogpt,ge2022long,yan2023temporally,yu2023language,weissenborn2019scaling,wu2021godiva,babaeizadeh2018stochastic,castrejon2019improved,denton2018stochastic,kondratyuk2024videopoet}, and implicit neural representations~\cite{skorokhodov2021stylegan,yu2021generating}.
Following the recent success of text-to-image Diffusion Models~\cite{rombach2022high, ramesh2022hierarchical, saharia2022photorealistic}, several works aim to achieve high-quality results for the text-to-video task. 
These works leverage diffusion process either in the pixel space~\cite{singer2022make,ge2023preserve,ho2022imagen} or latent space~\cite{blattmann2023align,zhou2022magicvideo, wang2023lavie, text2video-zero,luo2023videofusion,an2023latent,wang2023videofactory,wang2023lavie,xing2023simda,xing2023dynamicrafter, blattmann2023stable,gu2023reuse,gupta2023photorealistic,wang2023videolcm} or both~\cite{
zhang2023show1}. 
Reliably evaluating the above models remains a challenge. 
Current works primarily rely on FVD~\cite{unterthiner2019fvd} and human perceptual study. 
While a user study can reflect human preference more accurately, FVD serves as a more scalable evaluation protocol.
In this paper, we aim to better understand what aspects the FVD metric values more. 
Specifically, we analyze its sensitivity to the spatial versus temporal quality.

\paragraph{Evaluation metrics for image and video generation.} 
Many studies %have been done 
have focused on understanding and improving the evaluation metrics for image generation, such as Inception Score~\cite{salimans2016improved}, FID~\cite{kynkaanniemi2022role,parmar2022aliased,heusel2017gans,morozov2020self},  Perceptual Path Length~\cite{karras2019style}, and precision and recall~\cite{sajjadi2018assessing,kynkaanniemi2019improved}. 
% nash2021generating,morozov2020self,
Among them, FID is the most commonly adopted one, using the Inception-V3 feature extractor~\cite{Szegedy_2016_CVPR} trained on the ImageNet dataset~\cite{jia2009imagenet}. 
However, it can sometimes diverge from human judgment, especially on the out-of-domain datasets like human faces~\cite{NEURIPS2019_65699726,morozov2020self}.

To address the above issue, researchers have introduced several variants~\cite{Binkowski2018DemystifyingMG,sajjadi2018assessing,naeem2020reliable,kynkaanniemi2019improved} and performed analysis to understand FID~\cite{alfarra2022robustness,kynkaanniemi2022role,parmar2022aliased,borji2022pros}.
For instance, Kynk{\"a}{\"a}nniemi et al.~\cite{kynkaanniemi2022role} study the role of training data classes in the FID metric and advocate the use of the CLIP model as the feature extractor instead.
KID~\cite{Binkowski2018DemystifyingMG} is proposed to improve FID using the squared Maximum Mean Discrepancy (MMD) with a polynomial kernel. KID relaxes the Gaussian assumption in FID and requires fewer samples to compute.
Clean-FID~\cite{parmar2022aliased} shows that the aliasing issue caused by the preprocessing steps could significantly affect the FID scores. 
Similarly, Skorokhodov et al.~\cite{skorokhodov2021stylegan} studies the ``low-level'' preprocessing operations in FVD, such as resizing and frame sampling strategies. 
However, the analysis and improvement of the FVD are much less explored than those of FID.

\begin{figure}[t!]
    \centering
    \includegraphics[width=\linewidth]{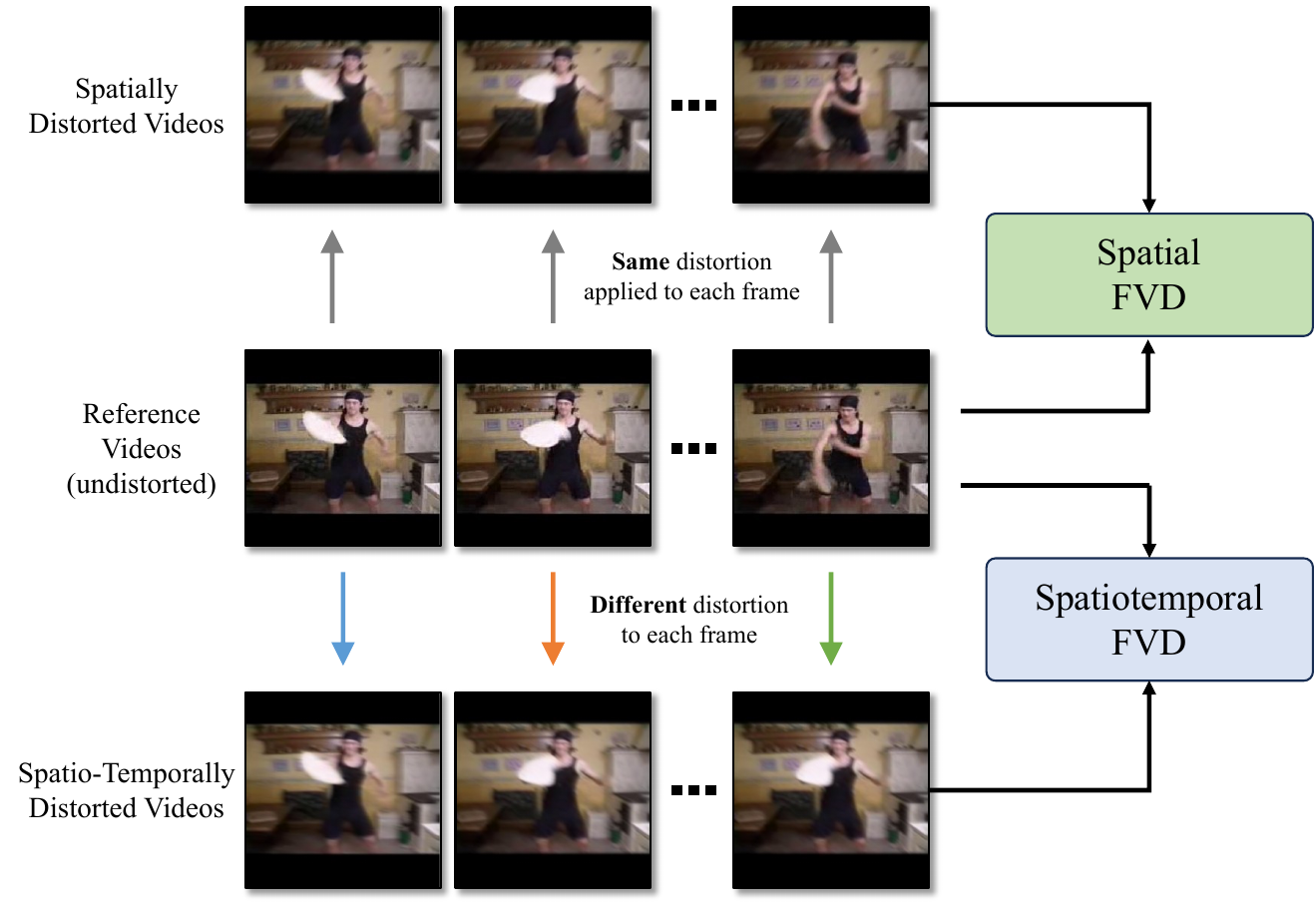}
    \caption{\textbf{Analyzing the FVD's sensitivity to temporal consistency.} We distort the same set of videos in spatial only or spatiotemporal manners so that the resulting videos have similar frame quality yet only differ in temporal quality. By comparing the FVD scores of the two distorted video sets, we aim at quantifying the temporal sensitivity of the metric.}
    % \vspace{-2mm}
    \label{fig:method_synthetic}
\end{figure}

\section{Quantifying the Temporal Sensitivity of FVD}

\begin{figure}
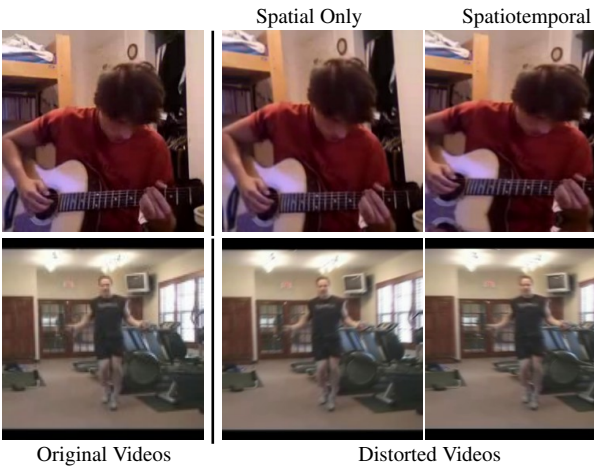

\setlength{\tabcolsep}{0.5pt}
\renewcommand{\arraystretch}{0.5}
\begin{tabularx}{\linewidth}{m{0.37\linewidth} c m{0.295\linewidth} c m{0.31\linewidth}} 
     & \hspace{0.6em} & 
    \footnotesize Spatial Only & \hspace{0.6em} &
    \footnotesize Spatiotemporal
\end{tabularx}

\begin{tabular}{c@{\hspace{3pt}}!{\vrule width 1pt}@{\hspace{3pt}}cc}
    \animategraphics[autoplay, loop, width=0.32\linewidth]{16}{videos/corruption/elastic_clean/}{0001}{0016} & 
    \animategraphics[autoplay, loop, width=0.32\linewidth]{16}{videos/corruption/elastic_spatial/}{0001}{0016} &
    \animategraphics[autoplay, loop, width=0.32\linewidth]{16}{videos/corruption/elastic_spatiotemporal/}{0001}{0016} 
\end{tabular}
% \begin{tabularx}{\linewidth}{m{0.05\linewidth} c m{0.23\linewidth} c } 
%     & \footnotesize Original Videos & \hspace{0.6em} & 
%     \footnotesize Elastic Transformation
% \end{tabularx}
\begin{tabular}{c@{\hspace{3pt}}!{\vrule width 1pt}@{\hspace{3pt}}cc}
    \animategraphics[autoplay, loop, width=0.32\linewidth]{16}{videos/corruption/motion_blur_clean/}{0001}{0016} & 
    \animategraphics[autoplay, loop, width=0.32\linewidth]{16}{videos/corruption/motion_blur_spatial/}{0001}{0016} &
    \animategraphics[autoplay, loop, width=0.32\linewidth]{16}{videos/corruption/motion_blur_spatiotemporal/}{0001}{0016} 
\end{tabular}
\begin{tabularx}{\linewidth}{m{0.05\linewidth} c m{0.29\linewidth} c } 
    & \footnotesize Original Videos & \hspace{0.6em} & 
    \footnotesize Distorted Videos
\end{tabularx}
\captionof{figure}{\textbf{Visualization of the spatial and spatiotemporal corruptions.} Both corruptions yield similar frame quality, while the spatiotemporal corruption induces additional temporal inconsistency in the video. By comparing the FVD of the spatiotemporal corruption with the spatial corruption, we analyze the temporal sensitivity of the metric. \emph{Best viewed with Acrobat Reader. 
%Click the images to play the video clips.
Please check our website for videos.}}
\vspace{-0.5cm}
\label{fig:vis_spatiotemporal}
\end{figure}

\begin{table*}[t!]
\setlength{\tabcolsep}{3pt}
\centering
\caption{\textbf{Analyzing FVD temporal sensitivity with video distortions.} 
We apply spatial only or spatiotemporal distortions to the videos. 
The two video sets share similar frame quality as assessed by FID. 
We thus use the FVD ratio to measure the temporal sensitivity of FVD.
}
\label{tab:spatiotemporal}
\begin{tabular}{llllllll}
\toprule
 Metric & Distortion & UCF-101 & Sky Time-lapse & FaceForensics & Taichi-HD & SSv2 & Kinectics-400 \\ \midrule
 \multirow{2}{*}{FID} & Spatial & 133.15 & 79.11 & 80.42 & 169.76 & 100.65 & 112.22 \\
  & Spatiotemporal & 133.69\tiny{$(+0.4\%)$} & 79.35\tiny{$(+0.3\%)$} & 79.57\tiny{$(-1.1\%)$} & 170.10\tiny{$(+0.2\%)$} & 100.62\tiny{$(-0.0\%)$} & 112.85\tiny{$(+0.6\%)$} \\ 
 \multirow{2}{*}{FVD} & Spatial & 1460.18 & 211.08 & 354.49 & 1016.78 & 594.68 & 996.71 \\ 
  & Spatiotemporal & 1705.27\tiny{$(+16.8\%)$} & 286.39\tiny{$(+35.7\%)$} & 367.35\tiny{$(+3.6\%)$} & 1201.35\tiny{$(+18.2\%)$} & 678.08\tiny{$(+14.0\%)$} & 1155.53\tiny{$(+15.9\%)$} \\
\bottomrule
\end{tabular}
\end{table*}

We examine the significance of temporal quality and consistency in FVD calculation. 
Recent studies suggest that models trained on the Kinetics datasets may not fully leverage the motion information~\cite{huang2018makes,liu2021no,Sevilla-Lara_2021_WACV,gberta_2021_ICML}, raising a similar question about whether the I3D features in FVD truly capture the motion quality of videos. 
One way to understand video motion \emph{vs}. content is to undermine one aspect through either spatial or temporal distortion. 
However, fully decoupling the two aspects is non-trivial~\cite{huang2018makes}. 
For example, poor frame quality would hinder the creation of a natural motion. 
As a result, previous approaches suggest creating videos by stitching real frames together, albeit in an incorrect order or from different videos~\cite{huang2018makes,Sevilla-Lara_2021_WACV,unterthiner2019fvd}. 

While this method is useful for analyzing video datasets, it may not be ideal for understanding a video generation metric. 
This is because generated videos rarely contain frames from irrelevant videos or arrange frames in incorrect order. 
% Nonetheless, the FVD study~\cite{unterthiner2019fvd} still observes a significant gap in FVD degradation 
% when introducing spatial and temporal corruptions. 
In contrast, we carefully design distortions that simulate real scenarios to quantify FVD's sensitivity to spatial and temporal video quality.

\myparagraph{Video distortion methods.} 
We illustrate our method for adding distortions in Figure~\ref{fig:method_synthetic}.
% As shown in Figure~\ref{fig:method_synthetic}, 
We apply two relevant distortions to the same set of real videos, aiming to synthesize videos with similar frame quality degradation but large differences in temporal quality. 
We employ conventional image distortion methods~\cite{Zhang_2018_CVPR,hendrycks2019robustness} including elastic transformation and motion blur. 
The elastic transformation locally stretches and contrasts the frames, while the motion blur averages image pixels along a specific direction of motion.  
We apply the same distortions to each frame to achieve a consistent frame quality drop.

To create spatiotemporal corruptions, as shown in Figure~\ref{fig:vis_spatiotemporal}, we apply randomly sampled elastic transformation parameters or blur kernels for each frame. 
This procedure allows us to introduce temporal inconsistency while producing similar frame quality to spatial corruptions. %For videos with spatial only distortions, we apply the same parameter for all the frames. 

\myparagraph{Experimental setups.} 
After distorting the videos using spatial only and spatiotemporal methods, we compute the FVD score of each video set with respect to the original videos. 
We apply distortion in five predefined corruption levels~\cite{hendrycks2019robustness} and compute the average. 
To verify that the two distorted sets have similar frame quality, we compute FID~\cite{parmar2022aliased} on the frames extracted from each set against the original image frames. %Since the FVD change is only affected by the temporal inconsistency introduced in the spatiotemporal distortion, 
Finally, we use the relative ratio between changes in FVD and FID to measure temporal sensitivity. 

\begin{figure}[t]
    \centering
    \includegraphics[width=\linewidth]{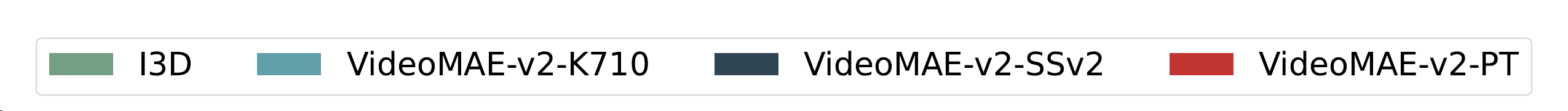}
    \begin{subfigure}{0.495\linewidth}
        \includegraphics[trim=0 0 0 0,clip,width=\linewidth]{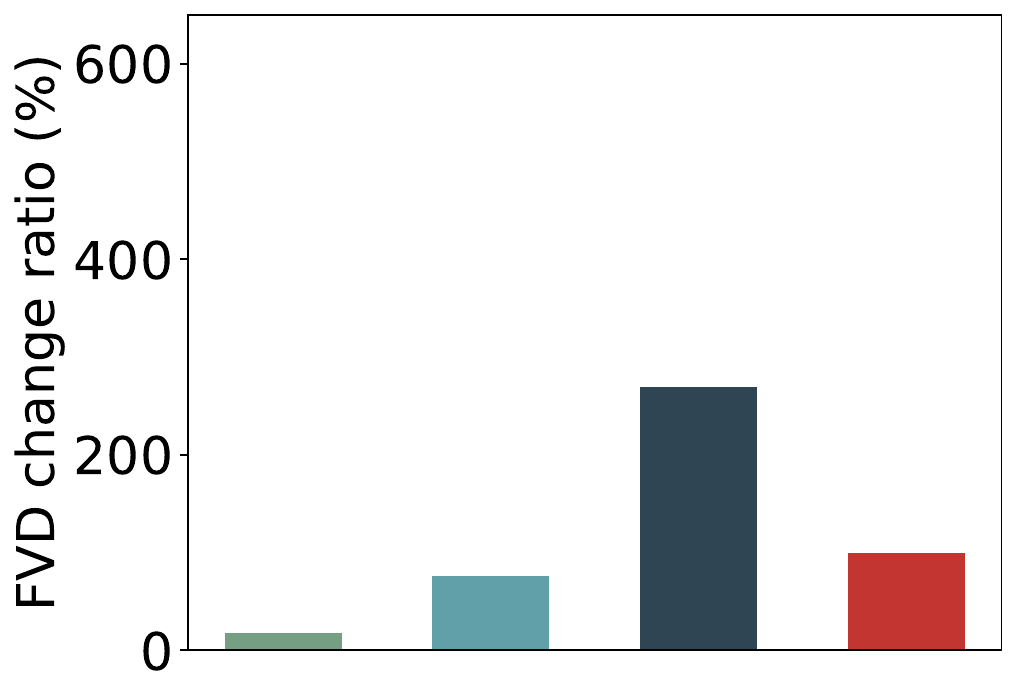}
        \caption{Motion Blur}
    \end{subfigure}
    \begin{subfigure}{0.465\linewidth}
        \includegraphics[trim=0 0 0 0,clip,width=\linewidth]{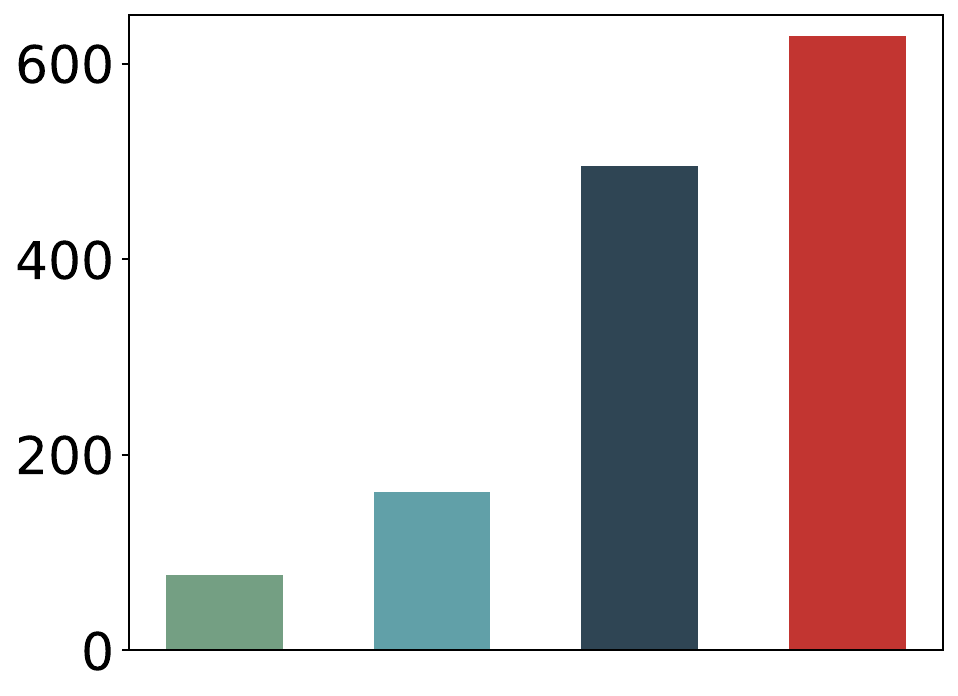}
        \caption{Elastic Transformation}
    \end{subfigure}
    \caption{\textbf{FVD sensitivity with different video feature extractors.}  We show that by substituting the I3D features with ones computed from the VideoMAE-v2 model, the temporal sensitivity can be significantly improved for both kinds of distortions. }
    \label{fig:videomae_st}
\end{figure}

We perform the experiments on several standard video datasets, including Kinetics-400~\cite{carreira2017quo}, Something-Something-v2~\cite{Goyal_2017_ICCV}, UCF-101~\cite{soomro2012ucf101}, Sky Time-lapse~\cite{xiong2018learning}, and FaceForensics~\cite{roessler2019faceforensicspp} datasets. 
Motivated by the previous finding that the unsupervised models trained on large datasets often produce more reliable features in FID~\cite{morozov2020self,kynkaanniemi2022role}, we also compute FVD using a self-supervised video model VideoMAE-v2~\cite{wang2023videomaev2}, which is trained on a mixed set of unlabeled datasets with the Masked Autoencoders (MAE) reconstruction objective~\cite{He_2022_CVPR}. 
Due to the large gap between the pertaining and downstream tasks, the MAE models are often further fine-tuned on the downstream tasks. 

\begin{figure}[t]
    \centering
    \includegraphics[width=\linewidth]{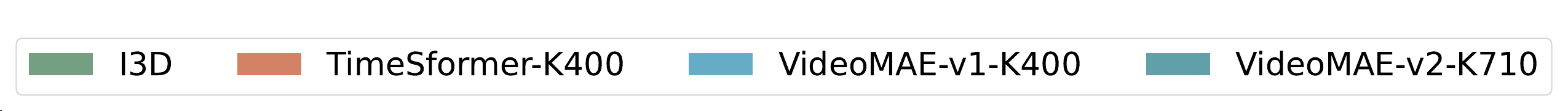}
    \begin{subfigure}{0.495\linewidth}
        \includegraphics[trim=0 0 0 0,clip,width=\linewidth]{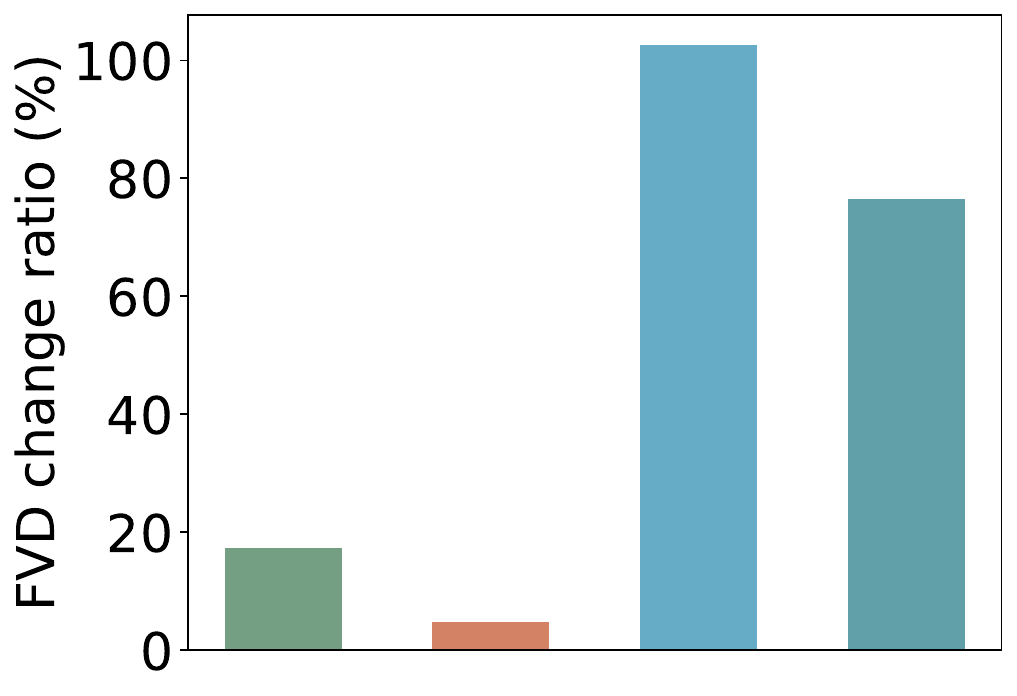}
        \caption{Motion Blur}
    \end{subfigure}
    \begin{subfigure}{0.465\linewidth}
        \includegraphics[trim=0 0 0 0,clip,width=\linewidth]{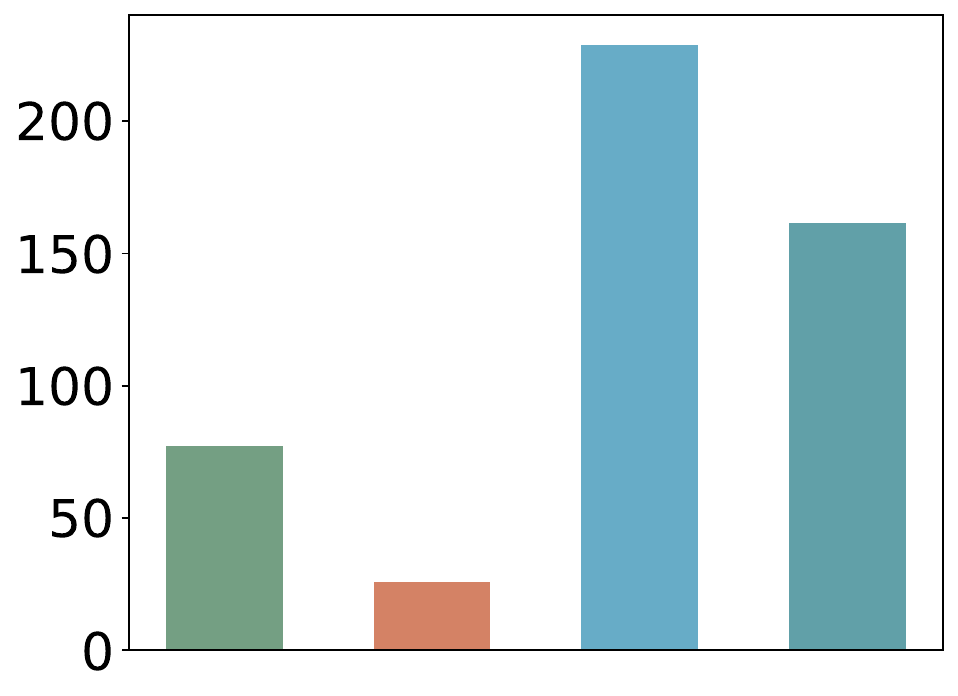}
        \caption{Elastic Transformation}
    \end{subfigure}
    \begin{tabular}{@{}lcccc@{}}
    \toprule
                     & \textcolor{i3d}{\rule{0.5cm}{0.2cm}}  & \textcolor{timesformerk400}{\rule{0.5cm}{0.2cm}} & \textcolor{videomaek400}{\rule{0.5cm}{0.2cm}} & \textcolor{videomaek710}{\rule{0.5cm}{0.2cm}}\\ \midrule
    ViT Architecture & \ding{55} & \ding{51} & \ding{51} & \ding{51} \\
    Self-Supervised Objective  & \ding{55} & \ding{55} & \ding{51} & \ding{51} \\
    Large-scale Training  & \ding{55} & \ding{55} & \ding{55} & \ding{51} \\ \bottomrule
    \end{tabular}
    
    \caption{\textbf{The origin of FVD sensitivity.} We show the temporal sensitivity achieved by using VideoMAE features is mainly attributed to the self-supervised objective.}
    \label{fig:videomae_origin}
\end{figure}

In our experiments, we explore a pretrained model \textit{VideoMAE-v2-PT} and two models fine-tuned on Kinetics-710 dataset~\cite{li2022uniformerv2} (\textit{VideoMAE-v2-K710}) and SSv2 dataset~\cite{Goyal_2017_ICCV} (\textit{VideoMAE-v2-SSv2}). 
More details about the dataset and experimental setups are included in Supplementary Material Section A.

\myparagraph{FVD temporal sensitivity.} 
We present FID and FVD scores obtained from the videos distorted spatially or spatiotemporally on different datasets in Table~\ref{tab:spatiotemporal}. 
We provide more results in Supplementary Material Section B. 
The minimal FID difference between the spatial and spatiotemporal distortion videos validates our claim that the two distorted video sets share similar frame quality.

Regarding the FVD scores of the spatially distorted videos, several datasets that are not in the distribution of the I3D training data, such as Sky Time-lapse, FaceForensics, and SSv2, generally yield much smaller FVD values compared to in-distribution datasets like Taichi-HD, UCF-101, and Kinectics-400. 
Given the same level of distortion is applied across different datasets, this suggests that FVD is less sensitive to distortion on the out-of-domain data. 
We inspect FVD's temporal sensitivity based on its increases induced by temporal inconsistency. Specifically, we compute the relative change of FVD between spatial and spatiotemporal corruptions. We find that FVD sometimes fails to detect the temporal quality decrease. 
For example, the temporal inconsistency in the FaceForensics dataset only raises FVD by $3\%$. 

To grasp the significance of the FVD increase due to temporal inconsistency, we compare it with the FVD values computed using VideoMAE-v2 models in Figure~\ref{fig:videomae_st}, where we report the average FVD scores across multiple datasets. 
When using the VideoMAE models to extract features, we notice a much more pronounced increase of FVD on the videos with temporal inconsistency. 
For example, when the elastic transformation is adopted to introduce temporal inconsistency, FVD with VideoMAE-v2-PT increases {\bf five} times more than the original FVD with the I3D model.

Compared to VideoMAE variants, the VideoMAE model fine-tuned on the SSv2 dataset consistently exhibits greater temporal sensitivity than the one fine-tuned on the K710 dataset, at least threefold. 
This difference can be attributed to the SSv2 dataset's emphasis on motion, where different videos share similar visual content, while differences only arise in fine-grained motion cues. 
Both \textit{VideoMAE-v2-SSv2} and \textit{VideoMAE-v2-PT} exhibit larger sensitivity to the temporal quality. Computing FVD with \textit{VideoMAE-v2-SSv2} features effectively captures mild temporal quality decrease induced by Motion Blur, whereas using \textit{VideoMAE-v2-PT} features proves more sensitive to temporal distortion introduced by Elastic Transformation. 
Overall, they all exhibit more sensitivity to temporal corruptions compared to FVD computed with the I3D model.

\myparagraph{Where does the content bias originate from?} 
Multiple factors could contribute to the increased temporal sensitivity when using features computed from the VideoMAE-v2 model. 
These factors may encompass the model architecture, training objectives, model capacity, and the dataset. 
To unravel these intricacies, we further delve into a comparative study with two other models, VideoMAE-v1~\cite{tong2022videomae} and TimeSFormer~\cite{gberta_2021_ICML} models.

The VideoMAE-v1 model uses a smaller ViT model size while sharing the same objective as the VideoMAE-v2 model, which helps us demystify the training scales. 
Due to limited computing resources, we cannot train the VideoMAE-v2 ViT model from scratch with the supervised objective. 
Instead, we train the smaller ViT with the size of VideoMAE-v1 using the recipe from TimeSFomer. 
Both models are trained on the Kinetics-400 dataset, sharing the same training dataset as the I3D model. 
For VideoMAE-v2, we use the \textit{VideoMAE-v2-K710} model, as it shares the most similar fine-tuning dataset, Kinetics-710, with other models. Note that it has the least temporal sensitivity among the three variants, as shown in Figure~\ref{fig:videomae_st}. 
We use it to make a fair comparison regarding the fine-tuning dataset.

We summarize the distinctions between these models in the table of Figure~\ref{fig:videomae_origin}. 
We perform our temporal sensitivity analysis and report the FVD ratio with different feature extractors in Figure~\ref{fig:videomae_origin}. 
The major improvement in the temporal sensitivity arises when comparing the VideoMAE-v1 and TimeSFormer models.
Based on these observations, we conclude that the self-supervised training objective contributes the most to mitigating the content bias.
\section{Probing the Perceptual Null Space in FVD}

\begin{figure}[t!]
    \centering
    \includegraphics[width=\linewidth]{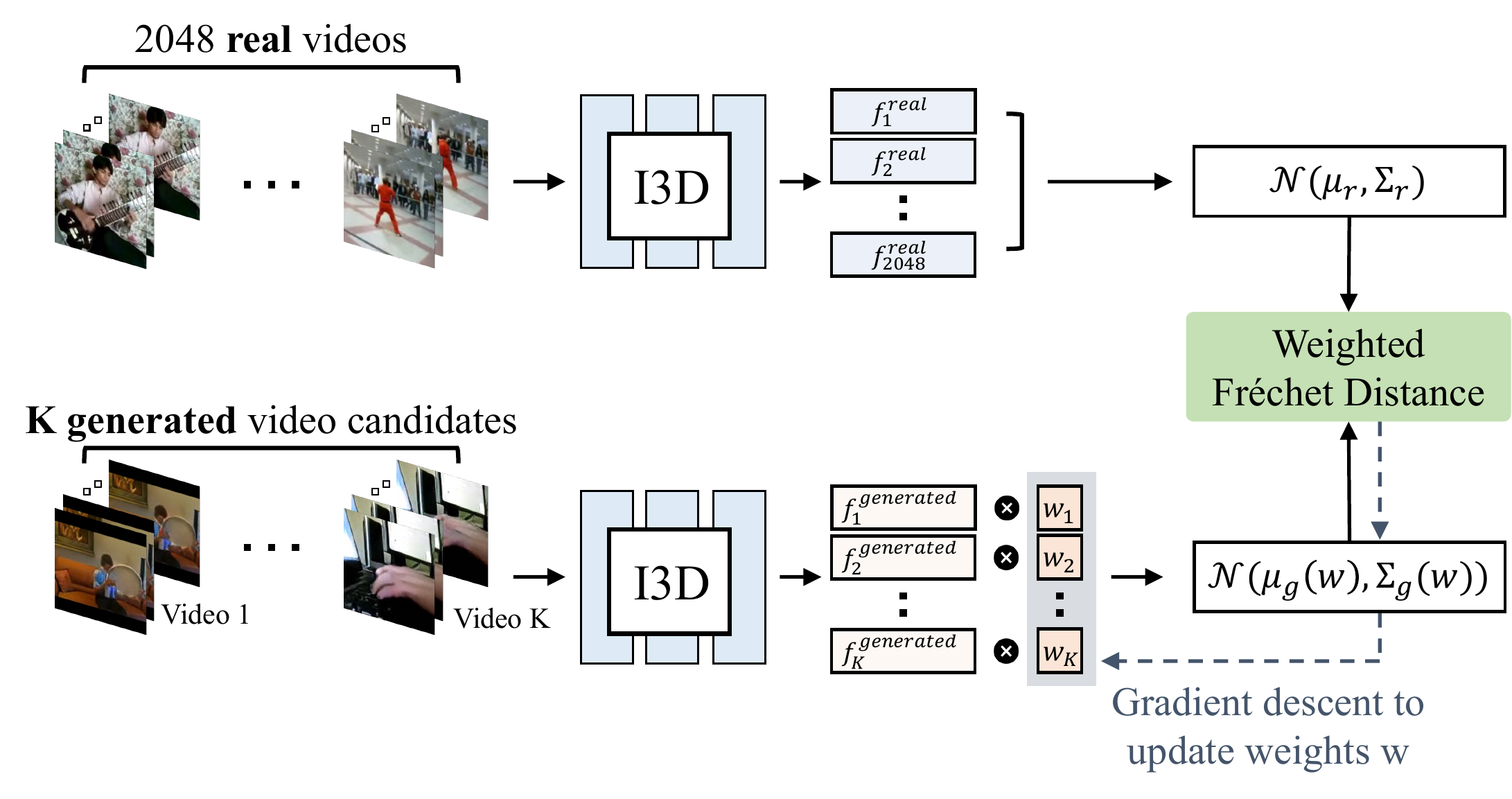}
    % \vspace{-8mm}
    \caption{
    \textbf{Probing the perceptual null space in FVD.} 
    We sample a $8\times$ larger set of fake videos and compute a weight for each candidate video by optimizing the weighted Frechet distance. 
    We then use the weights to sample $2,048$ videos to compute the final FVD score, which is used to measure the perceptual null space.}
    % \vspace{-2mm}
    \label{fig:method_null_space}
\end{figure}

Having observed FID's insensitivity to temporal quality in our synthetic experiments, a natural question arises: how does this bias impact practical evaluation? 
To address this question, we leverage an analysis method introduced to understand undesired behavior in FID~\cite{kynkaanniemi2022role}. 
This tool examines the \emph{perceptual null space},  where the quality of generated images remains similar while the FID score can be effectively adjusted. 
% In our system, the essential idea to exploit the perceptual null space is to generate a larger set of candidate
% videos from the same generation model and meticulously sample some videos from this larger set for evaluation to induce a decrease in FVD
In our scenario,  we generate a large set of candidate videos from the same model and carefully select a subset to lower the FVD score.

As the core assumption of the method, the samples synthesized by the same model should exhibit relatively similar visual quality. 
% However, we find that the quality may still vary for different generated videos in our case.
However, we observe that the quality may sometimes vary for different generative videos. 
Therefore, we further extend the analysis to understand the concept of temporal perceptual null space, where we hard-constrain the temporal quality of the generated videos to be the same by using frozen videos.

\myparagraph{Resampling method.} 
We adopt the resampling technique proposed by Kynkäänniemi et al.~\cite{kynkaanniemi2022role}.
Given a set of $K$, where $K>N$, generated videos, %it 
we assign a weight $w_k \in \mathbb{R}$ for each video, aiming to minimize the \emph{weighted FVD}. Given the weights,  the FVD defined in Equation~\ref{eq:fvd} is reformulated with the weighted mean $\mathbf{\mu}_g(\mathbf{w})=\frac{\sum_k{\exp{w_k}\mathbf{f}_k}}{\sum_k{\exp{w_k}}}$, and covariance  $\mathbf{\Sigma}_g(\mathbf{w}) = \frac{\sum_k\exp{w_k}(\mathbf{f}_k-\mathbf{\mu})(\mathbf{f}_k-\mathbf{\mu})^T}{\sum_k{\exp{w_k}}}$ as:
\begin{equation}
    \label{eq:weighted_fvd}
    \|\mathbf{\mu}_r-\mathbf{\mu}_g(\mathbf{w})\|^2_2 + \text{Tr}\left(\mathbf{\Sigma}_r + \mathbf{\Sigma}_g(\mathbf{w}) - 2 \left(\mathbf{\Sigma}_r\mathbf{\Sigma}_g(\mathbf{w})\right)^{\frac{1}{2}}\right).
\end{equation}
This \emph{weighted FVD} serves as the objective for optimizing $\mathbf{w}$. After the optimization process, the resampling is performed with the probability $\frac{\exp{w_k}}{\sum_k{\exp{w_k}}}$. $2,048$ videos are sampled from the candidate set to compute the new FVD score, which we denote as FVD\textsuperscript{*}.

\begin{figure}[t]
    \centering
    \begin{subfigure}[t]{0.495\linewidth}
        \includegraphics[trim=0 0 55 45,clip,width=\linewidth]{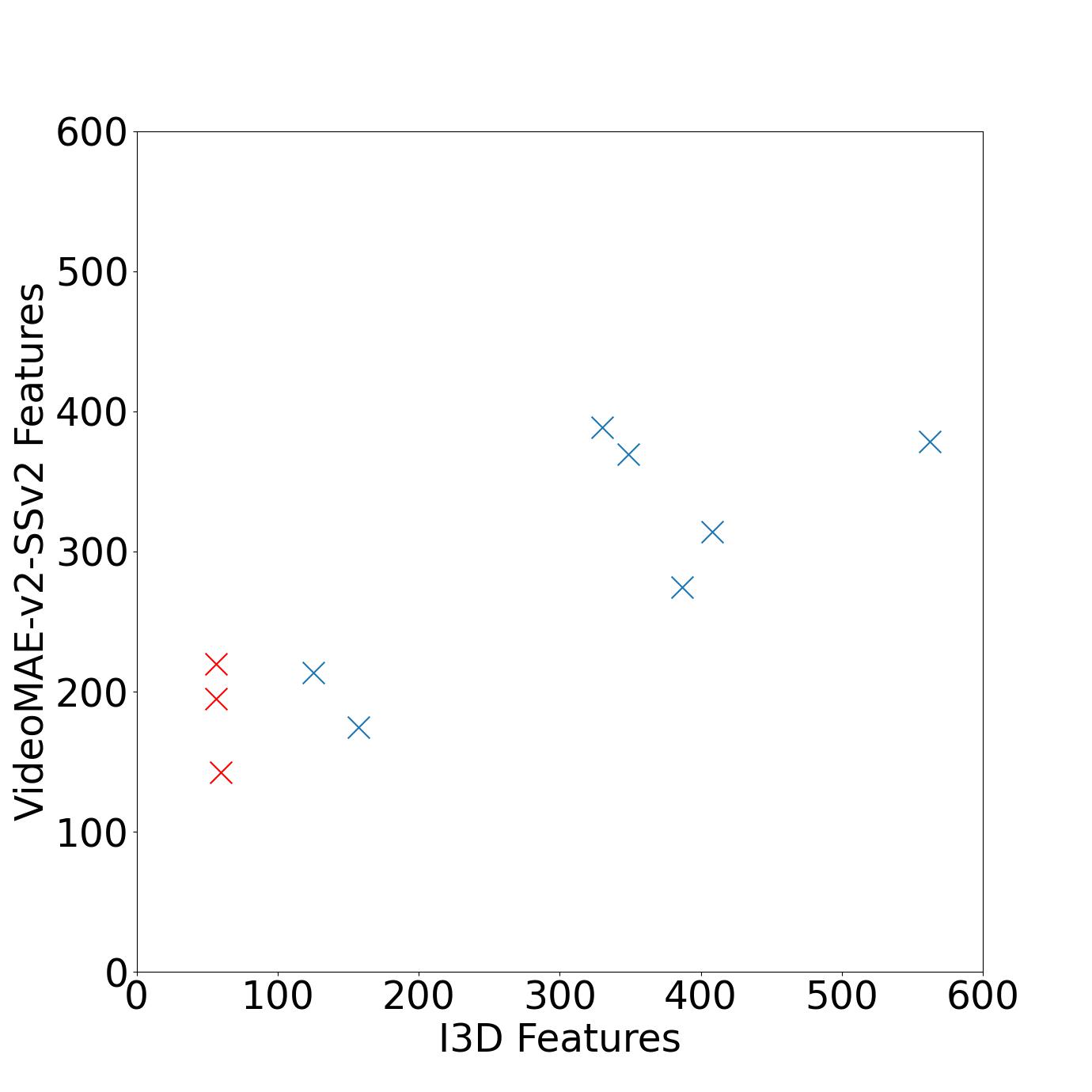}
        \centering
        \caption{FVD scores}
    \end{subfigure}
    \begin{subfigure}[t]{0.495\linewidth}
        \captionsetup{justification=centering}
        \includegraphics[trim=0 0 55 45,clip,width=\linewidth]{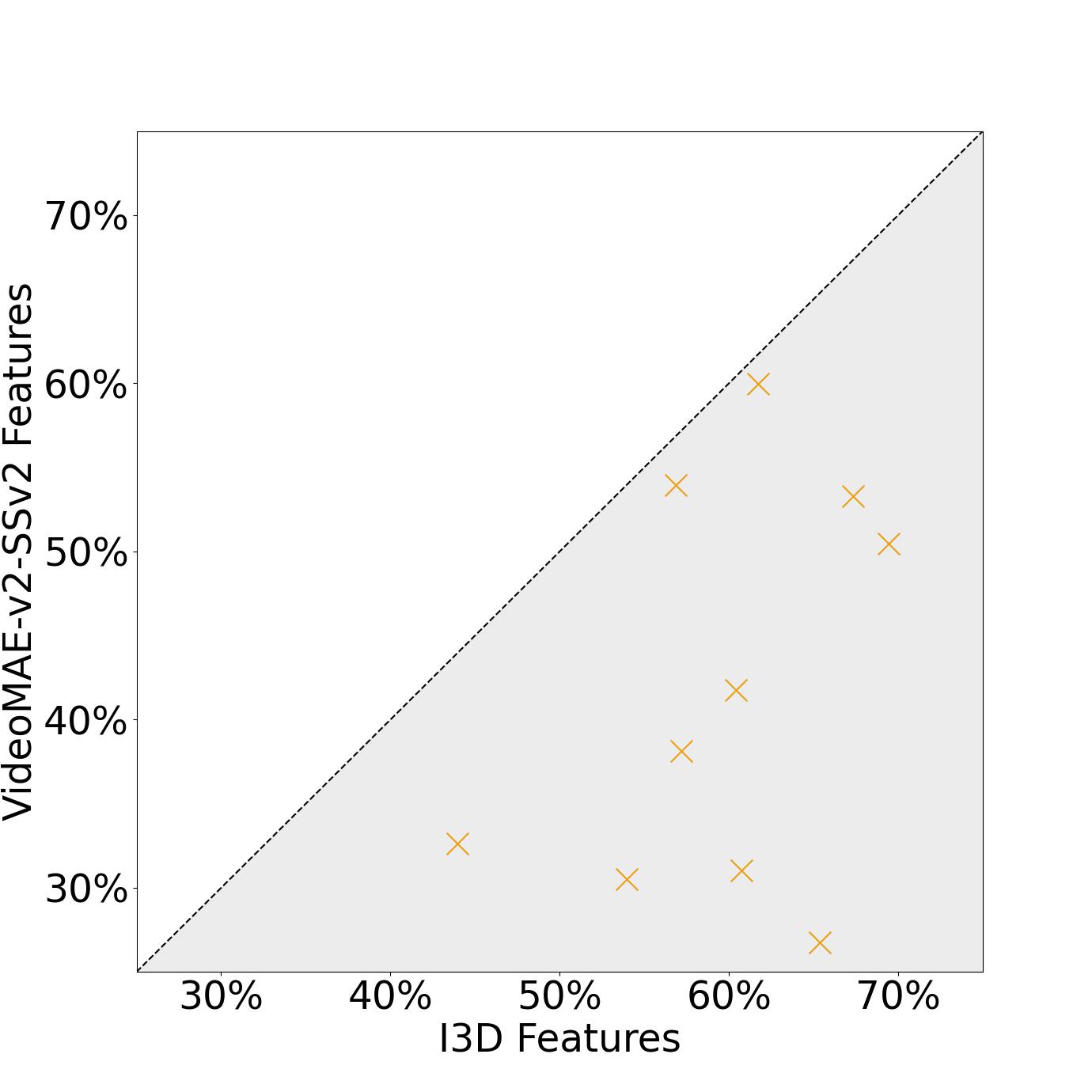}
        \caption{FVD decrease percentage \\ by resampling}
    \end{subfigure}
    \caption{\textbf{FVD decrease induced by resampling across different models and datasets.} We compare FVD computed from the VideoMAE-v2-SSv2 and I3D features. Each dot represents a video generation model trained on a specific dataset. (a) We notice a non-monotonic relationship between the FVD computed with I3D and VideoMAE-v2 features. For example, the models highlighted in red have similar FVD scores computed with I3D features but different scores when computed with VideoMAE-v2 features. (b) After resampling, the FVD with VideoMAE features generally decreases less than the FVD with I3D features, as most dots are located in the bottom right area (highlighted in grey).}
    \label{fig:resampling_scatter}
    \vspace{-0.3cm}
\end{figure}

\myparagraph{Experimental setups.} We experiment with several video generation models, including GANs~\cite{yu2021generating,skorokhodov2021stylegan}, Transformer~\cite{ge2022long}, and Diffusion models~\cite{yu2023video}. As these models are evaluated on different benchmarks, we also test with different datasets, including UCF-101~\cite{soomro2012ucf101}, Sky Time-lapse~\cite{xiong2018learning}, Taichi-HD~\cite{siarohin2019first}, and FaceForensics~\cite{roessler2019faceforensicspp}. We use the official scripts and checkpoints to sample the videos.
For each experiment, we generate a $8\times$ larger candidate set, i.e., $K=16,384$ videos. 
To optimize the weights $\mathbf{w}$, we use gradient descent with an initial learning rate of $0.01$ and a linear scheduler that decreases the learning rate by $0.1$ at every $100$ steps. 
We perform optimization for $300$ steps, at which point the weighted FVD scores often converge. 

We are especially interested in how much we can reduce the FVD score without improving the temporal quality. %To this end, we repeat the first frame of each generated video for $16$ times to form a \emph{frozen} video. 
To achieve this, we convert each generated video into a \emph{frozen} video by repeating its first frame 16 times. 
By doing so, we enforce all the videos to contain no motion so that the temporal quality cannot be improved through resampling.  Motivated by the temporal sensitivity presented by the VideoMAE-v2 models, we also perform the experiments with the \emph{VideoMAE-v2-SSv2} model as the feature extractor, where the resampling is done by optimizing its specific weights. Further details about the video generation models and experimental setups are included in Supplementary Material Section A.

% We focus on reducing the metric without enhancing temporal quality. To achieve this, we convert each generated video into a frozen video by repeating its first frame 16 times, ensuring no motion and preventing improvements in temporal quality through resampling. Driven by the temporal sensitivity of VideoMAE-v2 models, we also conduct experiments using the VideoMAE-v2-SSv2 model as the feature extractor, optimizing its specific weights. Further details on video generation models and experimental methods are in Supplementary Material Section A.

\begin{table*}[t]
\setlength{\tabcolsep}{3pt}
\centering
\caption{\textbf{Results of probing the temporal perceptual null space of FVD.} We report FVDs of normal and frozen generated videos by random sampling (FVD) and resampling to minimize weighted FVD (FVD\textsuperscript{*}). We color the FVD difference for better visualization: \textcolor{tablow}{$<20\%$}, \textcolor{tabmid}{$20\%-40\%$} and \textcolor{tabhigh}{$>40\%$}. The drop of FVD on the frozen generated videos indicates the volume of the null space where FVD can be reduced without generating a meaningful motion. The gray background indicates the samples where resampling frozen videos can obtain similar or even better FVD than the random generation results with motions.}
\label{tab:null_space}
\begin{tabular}{ll|lcl|lcl}
\toprule
& & \multicolumn{3}{c}{I3D Features} & \multicolumn{3}{c}{VideoMAE-v2-SSv2 Features} \\
Model      & Dataset    & FVD   & FVD\textsubscript{\scriptsize{w/o motion}}  & FVD\rlap{\textsubscript{\scriptsize{w/o motion}}}\textsuperscript{*} & FVD & FVD\textsubscript{\scriptsize{w/o motion}} & FVD\rlap{\textsubscript{\scriptsize{w/o motion}}}\textsuperscript{*} \\ \midrule
DIGAN~\cite{yu2021generating} & UCF-101~\cite{soomro2012ucf101} & 562.36 & 1303.13 & 715.96\scriptsize{$(\textcolor{tabhigh}{-45.1\%})$} & 378.19 & 951.59  & 859.57\scriptsize{$(\textcolor{tablow}{-9.7\%})$} \\ 
DIGAN~\cite{yu2021generating} & Sky Time-lapse~\cite{xiong2018learning} & \cellcolor[HTML]{DFDFDF}{157.13} & \cellcolor[HTML]{DFDFDF}{230.64} & \cellcolor[HTML]{DFDFDF}{115.55\scriptsize{$(\textcolor{tabhigh}{-49.9\%})$}} & 174.79 & 408.17  & 362.84\scriptsize{$(\textcolor{tablow}{-11.1\%})$} \\ 
DIGAN~\cite{yu2021generating} & Taichi-HD~\cite{siarohin2019first} & 132.26 & 461.79 & 276.88\scriptsize{$(\textcolor{tabhigh}{-40.0\%})$} & 313.84 & 578.61  & 523.20\scriptsize{$(\textcolor{tablow}{-9.6\%})$} \\ 
TATS~\cite{ge2022long} & UCF-101~\cite{soomro2012ucf101} & 329.92 & 1157.69 & 616.25\scriptsize{$(\textcolor{tabhigh}{-46.8\%})$} & 388.79 & 908.95  & 805.88\scriptsize{$(\textcolor{tablow}{-11.3\%})$} \\ 
TATS~\cite{ge2022long} & Sky Time-lapse~\cite{xiong2018learning} & \cellcolor[HTML]{DFDFDF}{125.62} & \cellcolor[HTML]{DFDFDF}{279.75} & \cellcolor[HTML]{DFDFDF}{126.32\scriptsize{$(\textcolor{tabhigh}{-54.8\%})$}} & 213.33 & 375.74  & 353.15\scriptsize{$(\textcolor{tablow}{-6.0\%})$} \\ 
TATS~\cite{ge2022long} & Taichi-HD~\cite{siarohin2019first} & 124.16 & 475.99 & 312.19\scriptsize{$(\textcolor{tabmid}{-34.4\%})$} & 274.81 & 587.31  & 530.86\scriptsize{$(\textcolor{tablow}{-9.6\%})$} \\ 
StyleGAN-V~\cite{skorokhodov2021stylegan} & Sky Time-lapse~\cite{xiong2018learning} & 56.63 & 206.56 & 104.27\scriptsize{$(\textcolor{tabhigh}{-49.5\%})$} & 219.85 & 503.22  & 456.24\scriptsize{$(\textcolor{tablow}{-9.3\%})$} \\ 
StyleGAN-V~\cite{skorokhodov2021stylegan} & FaceForensics~\cite{roessler2019faceforensicspp} & 56.22 & 353.79 & 242.04\scriptsize{$(\textcolor{tabmid}{-31.6\%})$} & 194.68 & 547.24  & 520.98\scriptsize{$(\textcolor{tablow}{-4.8\%})$} \\ 
PVDM~\cite{yu2023video} & UCF-101~\cite{soomro2012ucf101} & 348.81 & 1135.61 & 605.09\scriptsize{$(\textcolor{tabhigh}{-46.7\%})$} & 369.14 & 1032.90  & 898.48\scriptsize{$(\textcolor{tablow}{-13.0\%})$} \\ 
PVDM~\cite{yu2023video} & Sky Time-lapse~\cite{xiong2018learning} & 59.95 & 182.77 & 94.87\scriptsize{$(\textcolor{tabhigh}{-48.1\%})$} & 142.50 & 429.06  & 395.79\scriptsize{$(\textcolor{tablow}{-7.8\%})$} \\ 
\bottomrule
\end{tabular}
% \vspace{-0.3cm}
\end{table*}

\begin{figure}
\setlength{\tabcolsep}{0.5pt}
\renewcommand{\arraystretch}{0.5}
\begin{subfigure}{\linewidth}
\animategraphics[autoplay, loop, width=\linewidth]{16}{videos/null_space/digan_taichi_N_bottom32/}{0001}{0016}
% \begin{tabularx}{\linewidth}{m{0.1\linewidth} c} 
%    & Candidate videos with the smallest weights.
% \end{tabularx}
\caption{Candidate videos with the smallest weights.}
\end{subfigure}
\begin{subfigure}{\linewidth}
% \begin{tabular}{c}
\animategraphics[autoplay, loop, width=\linewidth]{16}{videos/null_space/digan_taichi_N_top32/}{0001}{0016}
% \end{tabular}
% \begin{tabularx}{\linewidth}{m{0.13\linewidth} c} 
%    & Candidate videos with the largest weights.
% \end{tabularx}
\caption{Candidate videos with the largest weights.}
\end{subfigure}
\vspace{-0.5cm}
\captionof{figure}{\textbf{Candidate videos with the largest and smallest weights.} We visualize the resampling results of DIGAN trained on the Taichi-HD dataset with the $32$ largest (most likely to select) and least (least likely to select) weights. We observe clear quality degradation in the samples with the smallest weights.  \emph{Best viewed with Acrobat Reader. 
%Click the images to play the video clips.
Please check the website for videos.}}
\vspace{-0.5cm}
\label{fig:vis_null_space}
\end{figure}

% \begin{figure}
% \setlength{\tabcolsep}{0.5pt}
% \renewcommand{\arraystretch}{0.5}
% \begin{subfigure}{\linewidth}
% \includegraphics[width=\linewidth]{videos/null_space/digan_taichi_N_bottom32/0001.jpg}
% \caption{Candidate videos with the smallest weights.}
% \end{subfigure}
% \begin{subfigure}{\linewidth}
% \includegraphics[width=\linewidth]{videos/null_space/digan_taichi_N_top32/0001.jpg}
% \caption{Candidate videos with the largest weights.}
% \end{subfigure}
% \vspace{-0.5cm}
% \captionof{figure}{\textbf{Candidate videos with the largest and smallest weights.} We visualize the resampling results of DIGAN trained on the Taichi-HD dataset with the $32$ largest (most likely to select) and least (least likely to select) weights. We observe clear quality degradation in the samples with the smallest weights.  \emph{Best viewed with Acrobat Reader. 
% %Click the images to play the video clips.
% Please check the website for videos.}}
% % \vspace{-0.5cm}
% \label{fig:vis_null_space}
% \end{figure}
\myparagraph{Observation on the resampling results.} We show FVD scores before resampling in Figure~\ref{fig:resampling_scatter} (a). We first observe a non-monotonic relationship between the FVD computed with I3D and VideoMAE-v2 features, potentially leading to different model rankings when using these two features. For FVD after resampling, we report the change ratio $\frac{\text{FVD}^*-\text{FVD}}{\text{FVD}}$ in Figure~\ref{fig:resampling_scatter} (b). It illustrates a significant drop ($40\% - 70\%$) in FVD computed with I3D features after resampling, while FVD computed with VideoMAE-v2 features experiences a more moderate drop ($25\% - 60\%$). 

We note clear quality differences after inspecting videos with the highest and lowest weights obtained from optimizing weighted FVD. An example of videos generated by DIGAN trained on the Taichi-HD dataset is shown in Figure~\ref{fig:vis_null_space}. We attribute the different observations from the original paper~\cite{kynkaanniemi2022role} to the unstable performance of the video generator. Since the video generator sometimes generates nonrealistic videos, resampling is beneficial in selecting a group of higher-quality videos, yielding smaller FVD scores.

\vspace{0.1cm}
\noindent\textbf{Temporal perceptual null space.} 
%Using frozen generated videos for resampling, we ensure that the temporal quality remains the same, as all videos contain no motion. 
Table~\ref{tab:null_space} shows extensive results, including the FVD of original videos with motions, the FVD of frozen videos, and the weighted FVD of frozen videos after resampling.  Despite the absence of motion in the generated videos, one can still reduce FVD by up to half by selectively choosing from the candidate videos when evaluating the Sky Time-lapse dataset. In the worst case, the same or even smaller FVD scores can be achieved compared with randomly selected generated videos \emph{with} motions. 

These findings highlight the pronounced content bias inherent in the FVD metric. Conversely, when computing the features for FVD using the VideoMAE-v2 model, which is sensitive to temporal quality, the gap significantly diminishes, and the FVD scores can hardly be decreased through resampling. This emphasizes that the FVD with VideoMAE-v2 has a much smaller temporal perceptual null space. More results are available in Supplementary Material Section B.

\section{Case Study: Long Video Generation}

% Our experiments have suggested that the motion in the generated videos is not valued enough by the FVD metric. 
Our experiments have revealed that FVD does not sufficiently account for motion in generated videos. We now dive into two case studies from previous works where FVD scores contradict human perception~\cite{ge2022long,skorokhodov2021stylegan,yu2021generating}. In both cases, the video generation models are trained on the Sky Time-lapse dataset, which is out-of-domain for the I3D model, and FVD has been shown not to perform well. In addition, both tasks generate longer videos than the standard $16$-frame setting, making the motion artifacts more perceptible to humans. Nevertheless, FVD fails to capture these motion artifacts in both experiments.

\myparagraph{Case study I~\cite{skorokhodov2021stylegan}.}
To synthesize long videos, the StyleGAN-v model~\cite{skorokhodov2021stylegan} employs convolutional layers with large reception fields to predict the parameters of the Fourier temporal encoding. We reproduce one of its baselines by substituting such temporal encoding with an LSTM layer. %As a visual result of generating videos with $128$ frames, 
For generating $128$-frame videos, the default StyleGAN-V synthesizes realistic motions (Figure \ref{fig:stylgan-v}a), whereas the baseline with continuous LSTM codes
% and $\sigma^z=16$ 
leads to videos with noticeable motion collapse (Figure \ref{fig:stylgan-v}b).

We follow the original study's evaluation protocol and compute the FVD metric by feeding all the 128 frames to the I3D model, termed as $\text{FVD}_{128}$. Note that the I3D model was initially trained on 64 frames, while the global average pooling and convolutional architecture allow it to be applied to any video length. We also compute $\text{FVD}_{128}$ with the VideoMAE. Since the VideoMAE uses a ViT with fixed-size positional encoding, we perform interpolation of positional encodings, similar to DINO~\cite{caron2021emerging}.

Contrary to the visual evidence, we observe the same trend as noted by the authors that $\text{FVD}_{128}$ computed using the I3D model is lower for the LSTM variant, compared to the original StyleGAN-v, as shown in Table \ref{tab:fvd_128}. Upon computing $\text{FVD}_{128}$ using VideoMAE-v2 features, both using SSv2 and K710, we have them to be in accordance with human preference, i.e., $\text{FVD}_{128}$ for the LSTM baseline is much worse compared to the original StyleGAN-v. 
%Since VideoMAE uses a ViT model with fixed size positional encoding, for computing $\text{FVD}_{128}$, we perform interpolation of positional encodings, similar to DINO~\cite{caron2021emerging}.

\begin{figure}[t!]
\setlength{\tabcolsep}{0.5pt}
\renewcommand{\arraystretch}{0.5}
\begin{subfigure}{\linewidth}
\animategraphics[autoplay, loop, width=\linewidth]{16}{videos/stylegan-v/default/}{0001}{0128}
\caption{Default StyleGAN-v.}
\end{subfigure}
% \begin{tabularx}{\linewidth}{m{0.33\linewidth} c} 
%     & Default StyleGAN-v.
% \end{tabularx}
\begin{subfigure}{\linewidth}
    \animategraphics[autoplay, loop, width=\linewidth]{16}{videos/stylegan-v/lstm/}{0001}{0128}
\caption{StyleGAN-v with LSTM motion codes.}
\end{subfigure}
% \vspace{-0.5cm}
\captionof{figure}{\textbf{Videos generated by StyelGAN-v and its LSTM variant.} The default StyleGAN-v synthesizes natural motions, while the variant with LSTM motion codes generates repeated patterns. \emph{Best viewed with Acrobat Reader. 
Please check the website for videos.}}
% \vspace{-0.3cm}
\label{fig:stylgan-v}
\end{figure}

% \begin{figure}[t!]
% \setlength{\tabcolsep}{0.5pt}
% \renewcommand{\arraystretch}{0.5}
% \begin{subfigure}{\linewidth}
%     \includegraphics[width=\linewidth]{videos/stylegan-v/default/0001.jpg}
% \caption{Default StyleGAN-v.}
% \end{subfigure}
% \begin{subfigure}{\linewidth}
%     \includegraphics[width=\linewidth]{videos/stylegan-v/lstm/0001.jpg}
% \caption{StyleGAN-v with LSTM motion codes.}
% \end{subfigure}
% \captionof{figure}{\textbf{Videos generated by StyelGAN-v and its LSTM variant.} The default StyleGAN-v synthesizes natural motions, while the variant with LSTM motion codes generates repeated patterns. \emph{Please check the website for videos.}
% }
% % \vspace{-0.3cm}
% \label{fig:stylgan-v}
% \end{figure}

% \begin{tabularx}{\linewidth}{m{0.33\linewidth} c} 
%     & Default StyleGAN-v.
% \end{tabularx}

\begin{table}[t!]
\setlength{\tabcolsep}{1pt}
\caption{StyleGAN-v model~\cite{skorokhodov2021stylegan} with LSTM as the motion codes trained on the Sky Time-lapse dataset generate collapsed motions, whereas FVD computed on the 128 frames favors the results. We show that computing FVD with VideoMAE-v2 features calibrates the conclusion.}
\label{tab:fvd_128}
\begin{tabular}{@{}llll@{}}
\toprule
Frame \#             & FVD Feature   & StylegGAN-v & w/ LSTM codes \\ \midrule
\multirow{3}{*}{16}  & I3D           & 120.11              & 136.65 \scriptsize{$(\textcolor{tablow}{+16.54\%})$} \\
                     & VideoMAE-SSv2 & 223.96              & 247.25\scriptsize{$(\textcolor{tablow}{+23.29\%})$} \\
                     & VideoMAE-K710 & 145.37              & 154.29\scriptsize{$(\textcolor{tablow}{+8.92\%})$} \\\hline
\multirow{3}{*}{128} & I3D           & 190.82              & 172.71\scriptsize{$(\textcolor{tabhigh}{-18.11\%})$}\\
                     & VideoMAE-SSv2 & 332.80              & 616.74\scriptsize{$(\textcolor{tablow}{+283.94\%})$}\\
                     & VideoMAE-K710 & 155.51              & 191.48\scriptsize{$(\textcolor{tablow}{+35.97\%})$}\\ \bottomrule
\end{tabular}
% \vspace{-0.3cm}
\end{table}

\myparagraph{Case study II~\cite{ge2022long,yu2021generating}.} Though trained on 16-frame video clips, DIGAN can be applied to generate longer videos by extrapolating the temporal encodings. However, in the previous study~\cite{ge2022long}, the authors have observed that motion artifacts in the form of repeated changes in the diagonal direction, while FVD computed on the $16$-frame chunks, favor the artifacts. 

Specifically, to evaluate the long video generation results, they compute FVD at strides of $k$ frames, $16$ frames at a time, for the entire video. We compute FVD on frames $0 \to 16$, $64 \to 82$, $128 \to 144$, and so on. The visualization of videos at $0 \to 16$ and $128 \to 144$ Figure~\ref{fig:digan} show more periodic artifacts at $128 \to 144$ compared to $0 \to 16$. In contrast, from Table~\ref{tab:fvd_parts}, we see that FVD computed using I3D features favors $128 \to 144$ frames, though FVD computed using VideoMAE features for both SSv2 and K710 follow human judgment.

Recent progress in photorealistic video generation using Diffusion models has enabled the creation of videos with extended durations ($\geq 100$ frames)~\cite{ge2023preserve, blattmann2023align, ho2022imagen, singer2022make}. 
%With the recent progress in photorealistic video generation with Diffusion models, it has been possible to generate videos with very long durations ($\geq 100$ frames)~\cite{ho2022imagen, blattmann2023align}. 
As a result, evaluating long video generation results has become increasingly important. However, according to the two real-world case studies above, FVD with I3D features does not reliably detect motion artifacts in long videos.

\begin{figure}[t!]
\setlength{\tabcolsep}{0.5pt}
\renewcommand{\arraystretch}{0.5}
\begin{subfigure}{\linewidth}
    \animategraphics[autoplay, loop, width=\linewidth]{4}{videos/DIGAN/0_16/}{0001}{0016}
    \caption{Frames 0 - 16.}
\end{subfigure}
% \begin{tabularx}{\linewidth}{m{0.35\linewidth} c} 
%     & Frames 0 - 16
% \end{tabularx}
\begin{subfigure}{\linewidth}
    \animategraphics[autoplay, loop, width=\linewidth]{4}{videos/DIGAN/128_144/}{0001}{0016}
    \caption{Frames 128 - 144.}
\end{subfigure}
% \begin{tabularx}{\linewidth}{m{0.33\linewidth} c} 
%     & Frames 128 - 144
% \end{tabularx}
\vspace{-0.5cm}
\captionof{figure}{\textbf{Videos generated by DIGAN at different extrapolated time steps.} The initial 16 frames generated by DIGAN exhibit natural motions, while the extrapolated frames contain periodic artifacts. \emph{Best viewed with Acrobat Reader. 
Please check the website for videos.}}
\vspace{-0.4cm}
\label{fig:digan}
\end{figure}

\begin{table}[t]
\caption{DIGAN~\cite{yu2021generating} trained on the Sky Time-lapse dataset with extrapolated time steps generate periodic artifacts, whereas the FVD metric favors the results. We show that computing FVD with VideoMAE-v2 features calibrates the conclusion.}
\centering
\label{tab:fvd_parts}
\begin{tabular}{@{}lll@{}}
\toprule
 FVD Feature   & Frames 0 - 16 & Frames 128 - 144 \\ \midrule
I3D         & 155.58              & 141.82\scriptsize{$(\textcolor{tabhigh}{-8.84\%})$}\\
VideoMAE-SSv2 & 133.37             & 150.61\scriptsize{$(\textcolor{tablow}{+12.9\%})$}\\
VideoMAE-K710 & 250.70            & 259.29\scriptsize{$(\textcolor{tablow}{+3.43\%})$}\\ \bottomrule
\end{tabular}
\vspace{-0.4cm}
\end{table}
\section{Discussion}
In this paper, we have studied the bias of the FVD on the frame quality. 
With experiments spanning from synthetic video distortion, to resampling video generation results, to investigating real-world examples, we have concluded that FVD is highly insensitive to the temporal quality and consistency of the generated videos. 
We have verified the hypothesis that the bias originates from the content-biased video features and show that self-supervised features can mitigate the issues in all the experiments. 
We hope our work will draw more attention to studying video generation evaluation and designing better evaluation metrics.

\myparagraph{Limitations.} 
Several critical aspects of FVD remain underexplored. 
For example, in addition to the longer time duration, existing methods also generate megapixel resolution videos~\cite{ge2023preserve, blattmann2023align, ho2022imagen, singer2022make}. 
However, to compute FVD (I3D or VideoMAE-v2), the video must be resized to a lower (e.g., $224\text{x}224$) resolution. 
In addition, many methods choose to generate videos not limited to the square aspect ratio, e.g., 16:9, while the FVD metric always requires a square video as the input. 
Computing FVD using VideoMAE-v2 is limited by the quadratic cost of attention layers, causing issues in evaluating longer video generation. Using efficient networks can reduce the cost, which we leave as future work.
% Anyway, we empirically observe that using half-precision has fewer effects than sampling real frames for computing FVD scores.

\myparagraph{Acknowledgment.} We thank Angjoo Kanazawa, Aleksander Holynski, Devi Parikh, and Yogesh Balaji for their early feedback and discussion. We thank Or Patashnik, Richard Zhang, Hanyu Wang, and Hadi Alzayer for their helpful comments, paper reading, and code reviewing. We thank Ivan Skorokhodov for his help with reproducing the StyleGAN-v experiments.
This work is partly supported by NSF grant No. IIS-239076, the Packard Fellowship, as well as NSF grants No. IIS-1910132 and IIS-2213335. 

\clearpage
{
    \small
    \bibliographystyle{ieeenat_fullname}
    \bibliography{main}

\begin{thebibliography}{96}
\providecommand{\natexlab}[1]{#1}
\providecommand{\url}[1]{\texttt{#1}}
\expandafter\ifx\csname urlstyle\endcsname\relax
  \providecommand{\doi}[1]{doi: #1}\else
  \providecommand{\doi}{doi: \begingroup \urlstyle{rm}\Url}\fi

\bibitem[Alfarra et~al.(2022)Alfarra, P{\'e}rez, Fr{\"u}hst{\"u}ck, Torr, Wonka, and Ghanem]{alfarra2022robustness}
Motasem Alfarra, Juan~C P{\'e}rez, Anna Fr{\"u}hst{\"u}ck, Philip~HS Torr, Peter Wonka, and Bernard Ghanem.
\newblock On the robustness of quality measures for gans.
\newblock In \emph{ECCV}, 2022.

\bibitem[An et~al.(2023)An, Zhang, Yang, Gupta, Huang, Luo, and Yin]{an2023latent}
Jie An, Songyang Zhang, Harry Yang, Sonal Gupta, Jia-Bin Huang, Jiebo Luo, and Xi Yin.
\newblock Latent-shift: Latent diffusion with temporal shift for efficient text-to-video generation.
\newblock \emph{arXiv preprint arXiv:2304.08477}, 2023.

\bibitem[Babaeizadeh et~al.(2018)Babaeizadeh, Finn, Erhan, Campbell, and Levine]{babaeizadeh2018stochastic}
Mohammad Babaeizadeh, Chelsea Finn, Dumitru Erhan, Roy~H Campbell, and Sergey Levine.
\newblock Stochastic variational video prediction.
\newblock In \emph{ICLR}, 2018.

\bibitem[Bain et~al.(2021)Bain, Nagrani, Varol, and Zisserman]{bain2021frozen}
Max Bain, Arsha Nagrani, G{\"u}l Varol, and Andrew Zisserman.
\newblock Frozen in time: A joint video and image encoder for end-to-end retrieval.
\newblock In \emph{ICCV}, pages 1728--1738, 2021.

\bibitem[Bar-Tal et~al.(2024)Bar-Tal, Chefer, Tov, Herrmann, Paiss, Zada, Ephrat, Hur, Liu, Raj, Li, Rubinstein, Michaeli, Wang, Sun, Dekel, and Mosseri]{bartal2024lumiere}
Omer Bar-Tal, Hila Chefer, Omer Tov, Charles Herrmann, Roni Paiss, Shiran Zada, Ariel Ephrat, Junhwa Hur, Guanghui Liu, Amit Raj, Yuanzhen Li, Michael Rubinstein, Tomer Michaeli, Oliver Wang, Deqing Sun, Tali Dekel, and Inbar Mosseri.
\newblock Lumiere: A space-time diffusion model for video generation, 2024.

\bibitem[Bertasius et~al.(2021)Bertasius, Wang, and Torresani]{gberta_2021_ICML}
Gedas Bertasius, Heng Wang, and Lorenzo Torresani.
\newblock Is space-time attention all you need for video understanding?
\newblock In \emph{ICML}, 2021.

\bibitem[Binkowski et~al.(2018)Binkowski, Sutherland, Arbel, and Gretton]{Binkowski2018DemystifyingMG}
Mikolaj Binkowski, Danica~J. Sutherland, Michal Arbel, and Arthur Gretton.
\newblock Demystifying mmd gans.
\newblock In \emph{ICLR}, 2018.

\bibitem[Blattmann et~al.(2023{\natexlab{a}})Blattmann, Dockhorn, Kulal, Mendelevitch, Kilian, Lorenz, Levi, English, Voleti, Letts, Jampani, and Rombach]{blattmann2023stable}
Andreas Blattmann, Tim Dockhorn, Sumith Kulal, Daniel Mendelevitch, Maciej Kilian, Dominik Lorenz, Yam Levi, Zion English, Vikram Voleti, Adam Letts, Varun Jampani, and Robin Rombach.
\newblock Stable video diffusion: Scaling latent video diffusion models to large datasets, 2023{\natexlab{a}}.

\bibitem[Blattmann et~al.(2023{\natexlab{b}})Blattmann, Rombach, Ling, Dockhorn, Kim, Fidler, and Kreis]{blattmann2023align}
Andreas Blattmann, Robin Rombach, Huan Ling, Tim Dockhorn, Seung~Wook Kim, Sanja Fidler, and Karsten Kreis.
\newblock Align your latents: High-resolution video synthesis with latent diffusion models.
\newblock In \emph{CVPR}, 2023{\natexlab{b}}.

\bibitem[Borji(2022)]{borji2022pros}
Ali Borji.
\newblock Pros and cons of gan evaluation measures: New developments.
\newblock \emph{Computer Vision and Image Understanding}, 215:\penalty0 103329, 2022.

\bibitem[Brooks et~al.(2022)Brooks, Hellsten, Aittala, Wang, Aila, Lehtinen, Liu, Efros, and Karras]{brooks2022generating}
Tim Brooks, Janne Hellsten, Miika Aittala, Ting-Chun Wang, Timo Aila, Jaakko Lehtinen, Ming-Yu Liu, Alexei Efros, and Tero Karras.
\newblock Generating long videos of dynamic scenes.
\newblock \emph{NeurIPS}, 35:\penalty0 31769--31781, 2022.

\bibitem[Brooks et~al.(2024)Brooks, Peebles, Holmes, DePue, Guo, Jing, Schnurr, Taylor, Luhman, Luhman, Ng, Wang, and Ramesh]{videoworldsimulators2024}
Tim Brooks, Bill Peebles, Connor Holmes, Will DePue, Yufei Guo, Li Jing, David Schnurr, Joe Taylor, Troy Luhman, Eric Luhman, Clarence Ng, Ricky Wang, and Aditya Ramesh.
\newblock Video generation models as world simulators, 2024.

\bibitem[Caron et~al.(2021)Caron, Touvron, Misra, J{\'e}gou, Mairal, Bojanowski, and Joulin]{caron2021emerging}
Mathilde Caron, Hugo Touvron, Ishan Misra, Herv{\'e} J{\'e}gou, Julien Mairal, Piotr Bojanowski, and Armand Joulin.
\newblock Emerging properties in self-supervised vision transformers.
\newblock In \emph{ICCV}, 2021.

\bibitem[Carreira and Zisserman(2017)]{carreira2017quo}
Joao Carreira and Andrew Zisserman.
\newblock Quo vadis, action recognition? a new model and the kinetics dataset.
\newblock In \emph{CVPR}, pages 6299--6308, 2017.

\bibitem[Castrejon et~al.(2019)Castrejon, Ballas, and Courville]{castrejon2019improved}
Lluis Castrejon, Nicolas Ballas, and Aaron Courville.
\newblock Improved conditional vrnns for video prediction.
\newblock In \emph{ICCV}, pages 7608--7617, 2019.

\bibitem[Choi et~al.(2019)Choi, Gao, Messou, and Huang]{choi2019can}
Jinwoo Choi, Chen Gao, Joseph~CE Messou, and Jia-Bin Huang.
\newblock Why can't i dance in the mall? learning to mitigate scene bias in action recognition.
\newblock In \emph{NeurIPS}, 2019.

\bibitem[Deng et~al.(2009)Deng, Dong, Socher, Li, Li, and Fei-Fei]{jia2009imagenet}
Jia Deng, Wei Dong, Richard Socher, Li-Jia Li, Kai Li, and Li Fei-Fei.
\newblock Imagenet: A large-scale hierarchical image database.
\newblock In \emph{CVPR}, pages 248--255, 2009.

\bibitem[Denton and Fergus(2018)]{denton2018stochastic}
Emily Denton and Rob Fergus.
\newblock Stochastic video generation with a learned prior.
\newblock In \emph{ICML}, pages 1174--1183. PMLR, 2018.

\bibitem[Dowson and Landau(1982)]{dowson1982frechet}
DC Dowson and BV Landau.
\newblock The fr{\'e}chet distance between multivariate normal distributions.
\newblock \emph{Journal of multivariate analysis}, 12\penalty0 (3):\penalty0 450--455, 1982.

\bibitem[Ge et~al.(2022)Ge, Hayes, Yang, Yin, Pang, Jacobs, Huang, and Parikh]{ge2022long}
Songwei Ge, Thomas Hayes, Harry Yang, Xi Yin, Guan Pang, David Jacobs, Jia-Bin Huang, and Devi Parikh.
\newblock Long video generation with time-agnostic vqgan and time-sensitive transformer.
\newblock In \emph{ECCV}, 2022.

\bibitem[Ge et~al.(2023)Ge, Nah, Liu, Poon, Tao, Catanzaro, Jacobs, Huang, Liu, and Balaji]{ge2023preserve}
Songwei Ge, Seungjun Nah, Guilin Liu, Tyler Poon, Andrew Tao, Bryan Catanzaro, David Jacobs, Jia-Bin Huang, Ming-Yu Liu, and Yogesh Balaji.
\newblock Preserve your own correlation: A noise prior for video diffusion models.
\newblock In \emph{ICCV}, 2023.

\bibitem[Girdhar et~al.(2023)Girdhar, Singh, Brown, Duval, Azadi, Rambhatla, Shah, Yin, Parikh, and Misra]{girdhar2023emu}
Rohit Girdhar, Mannat Singh, Andrew Brown, Quentin Duval, Samaneh Azadi, Sai~Saketh Rambhatla, Akbar Shah, Xi Yin, Devi Parikh, and Ishan Misra.
\newblock Emu video: Factorizing text-to-video generation by explicit image conditioning, 2023.

\bibitem[Goyal et~al.(2017)Goyal, Ebrahimi~Kahou, Michalski, Materzynska, Westphal, Kim, Haenel, Fruend, Yianilos, Mueller-Freitag, Hoppe, Thurau, Bax, and Memisevic]{Goyal_2017_ICCV}
Raghav Goyal, Samira Ebrahimi~Kahou, Vincent Michalski, Joanna Materzynska, Susanne Westphal, Heuna Kim, Valentin Haenel, Ingo Fruend, Peter Yianilos, Moritz Mueller-Freitag, Florian Hoppe, Christian Thurau, Ingo Bax, and Roland Memisevic.
\newblock The "something something" video database for learning and evaluating visual common sense.
\newblock In \emph{ICCV}, 2017.

\bibitem[Gu et~al.(2023)Gu, Wang, Zhao, Lu, Zhang, Wu, Xu, Zhang, Jiang, and Xu]{gu2023reuse}
Jiaxi Gu, Shicong Wang, Haoyu Zhao, Tianyi Lu, Xing Zhang, Zuxuan Wu, Songcen Xu, Wei Zhang, Yu-Gang Jiang, and Hang Xu.
\newblock Reuse and diffuse: Iterative denoising for text-to-video generation, 2023.

\bibitem[Gupta et~al.(2023)Gupta, Yu, Sohn, Gu, Hahn, Fei-Fei, Essa, Jiang, and Lezama]{gupta2023photorealistic}
Agrim Gupta, Lijun Yu, Kihyuk Sohn, Xiuye Gu, Meera Hahn, Li Fei-Fei, Irfan Essa, Lu Jiang, and José Lezama.
\newblock Photorealistic video generation with diffusion models, 2023.

\bibitem[He et~al.(2022)He, Chen, Xie, Li, Doll\'ar, and Girshick]{He_2022_CVPR}
Kaiming He, Xinlei Chen, Saining Xie, Yanghao Li, Piotr Doll\'ar, and Ross Girshick.
\newblock Masked autoencoders are scalable vision learners.
\newblock In \emph{CVPR}, pages 16000--16009, 2022.

\bibitem[Hendrycks and Dietterich(2019)]{hendrycks2019robustness}
Dan Hendrycks and Thomas Dietterich.
\newblock Benchmarking neural network robustness to common corruptions and perturbations.
\newblock In \emph{ICLR}, 2019.

\bibitem[Heusel et~al.(2017)Heusel, Ramsauer, Unterthiner, Nessler, and Hochreiter]{heusel2017gans}
Martin Heusel, Hubert Ramsauer, Thomas Unterthiner, Bernhard Nessler, and Sepp Hochreiter.
\newblock Gans trained by a two time-scale update rule converge to a local nash equilibrium.
\newblock In \emph{NeurIPS}, 2017.

\bibitem[Ho et~al.(2020)Ho, Jain, and Abbeel]{ho2020denoising}
Jonathan Ho, Ajay Jain, and Pieter Abbeel.
\newblock Denoising diffusion probabilistic models.
\newblock In \emph{NeurIPS}, 2020.

\bibitem[Ho et~al.(2022{\natexlab{a}})Ho, Chan, Saharia, Whang, Gao, Gritsenko, Kingma, Poole, Norouzi, Fleet, et~al.]{ho2022imagen}
Jonathan Ho, William Chan, Chitwan Saharia, Jay Whang, Ruiqi Gao, Alexey Gritsenko, Diederik~P Kingma, Ben Poole, Mohammad Norouzi, David~J Fleet, et~al.
\newblock Imagen video: High definition video generation with diffusion models.
\newblock \emph{arXiv preprint arXiv:2210.02303}, 2022{\natexlab{a}}.

\bibitem[Ho et~al.(2022{\natexlab{b}})Ho, Salimans, Gritsenko, Chan, Norouzi, and Fleet]{ho2022video}
Jonathan Ho, Tim Salimans, Alexey Gritsenko, William Chan, Mohammad Norouzi, and David~J Fleet.
\newblock Video diffusion models.
\newblock \emph{arXiv preprint arXiv:2204.03458}, 2022{\natexlab{b}}.

\bibitem[Huang et~al.(2018)Huang, Ramanathan, Mahajan, Torresani, Paluri, Fei-Fei, and Niebles]{huang2018makes}
De-An Huang, Vignesh Ramanathan, Dhruv Mahajan, Lorenzo Torresani, Manohar Paluri, Li Fei-Fei, and Juan~Carlos Niebles.
\newblock What makes a video a video: Analyzing temporal information in video understanding models and datasets.
\newblock In \emph{CVPR}, 2018.

\bibitem[Huang et~al.(2023)Huang, He, Yu, Zhang, Si, Jiang, Zhang, Wu, Jin, Chanpaisit, Wang, Chen, Wang, Lin, Qiao, and Liu]{huang2023vbench}
Ziqi Huang, Yinan He, Jiashuo Yu, Fan Zhang, Chenyang Si, Yuming Jiang, Yuanhan Zhang, Tianxing Wu, Qingyang Jin, Nattapol Chanpaisit, Yaohui Wang, Xinyuan Chen, Limin Wang, Dahua Lin, Yu Qiao, and Ziwei Liu.
\newblock Vbench: Comprehensive benchmark suite for video generative models, 2023.

\bibitem[Jang et~al.(2018)Jang, Kim, and Song]{pmlr-v80-jang18a}
Yunseok Jang, Gunhee Kim, and Yale Song.
\newblock Video prediction with appearance and motion conditions.
\newblock In \emph{ICML}, pages 2225--2234, 2018.

\bibitem[Karras et~al.(2019)Karras, Laine, and Aila]{karras2019style}
Tero Karras, Samuli Laine, and Timo Aila.
\newblock A style-based generator architecture for generative adversarial networks.
\newblock In \emph{CVPR}, 2019.

\bibitem[Khachatryan et~al.(2023)Khachatryan, Movsisyan, Tadevosyan, Henschel, Wang, Navasardyan, and Shi]{text2video-zero}
Levon Khachatryan, Andranik Movsisyan, Vahram Tadevosyan, Roberto Henschel, Zhangyang Wang, Shant Navasardyan, and Humphrey Shi.
\newblock Text2video-zero: Text-to-image diffusion models are zero-shot video generators.
\newblock In \emph{CVPR}, 2023.

\bibitem[Kondratyuk et~al.(2024)Kondratyuk, Yu, Gu, Lezama, Huang, Hornung, Adam, Akbari, Alon, Birodkar, Cheng, Chiu, Dillon, Essa, Gupta, Hahn, Hauth, Hendon, Martinez, Minnen, Ross, Schindler, Sirotenko, Sohn, Somandepalli, Wang, Yan, Yang, Yang, Seybold, and Jiang]{kondratyuk2024videopoet}
Dan Kondratyuk, Lijun Yu, Xiuye Gu, José Lezama, Jonathan Huang, Rachel Hornung, Hartwig Adam, Hassan Akbari, Yair Alon, Vighnesh Birodkar, Yong Cheng, Ming-Chang Chiu, Josh Dillon, Irfan Essa, Agrim Gupta, Meera Hahn, Anja Hauth, David Hendon, Alonso Martinez, David Minnen, David Ross, Grant Schindler, Mikhail Sirotenko, Kihyuk Sohn, Krishna Somandepalli, Huisheng Wang, Jimmy Yan, Ming-Hsuan Yang, Xuan Yang, Bryan Seybold, and Lu Jiang.
\newblock Videopoet: A large language model for zero-shot video generation, 2024.

\bibitem[Kynk{\"a}{\"a}nniemi et~al.(2019)Kynk{\"a}{\"a}nniemi, Karras, Laine, Lehtinen, and Aila]{kynkaanniemi2019improved}
Tuomas Kynk{\"a}{\"a}nniemi, Tero Karras, Samuli Laine, Jaakko Lehtinen, and Timo Aila.
\newblock Improved precision and recall metric for assessing generative models.
\newblock In \emph{NeurIPS}, 2019.

\bibitem[Kynk{\"a}{\"a}nniemi et~al.(2022)Kynk{\"a}{\"a}nniemi, Karras, Aittala, Aila, and Lehtinen]{kynkaanniemi2022role}
Tuomas Kynk{\"a}{\"a}nniemi, Tero Karras, Miika Aittala, Timo Aila, and Jaakko Lehtinen.
\newblock The role of imagenet classes in fr{\'e}chet inception distance.
\newblock In \emph{ICLR}, 2022.

\bibitem[Li et~al.(2019)Li, Jiang, and Jiang]{li2019quality}
Dingquan Li, Tingting Jiang, and Ming Jiang.
\newblock Quality assessment of in-the-wild videos.
\newblock In \emph{Proceedings of the 27th ACM International Conference on Multimedia}, pages 2351--2359, 2019.

\bibitem[Li et~al.(2022)Li, Wang, He, Li, Wang, Wang, and Qiao]{li2022uniformerv2}
Kunchang Li, Yali Wang, Yinan He, Yizhuo Li, Yi Wang, Limin Wang, and Yu Qiao.
\newblock Uniformerv2: Spatiotemporal learning by arming image vits with video uniformer, 2022.

\bibitem[Li et~al.(2018)Li, Li, and Vasconcelos]{li2018resound}
Yingwei Li, Yi Li, and Nuno Vasconcelos.
\newblock Resound: Towards action recognition without representation bias.
\newblock In \emph{ECCV}, 2018.

\bibitem[Liu et~al.(2021)Liu, Pintea, Nejadasl, Booij, and Van~Gemert]{liu2021no}
Xin Liu, Silvia~L Pintea, Fatemeh~Karimi Nejadasl, Olaf Booij, and Jan~C Van~Gemert.
\newblock No frame left behind: Full video action recognition.
\newblock In \emph{CVPR}, 2021.

\bibitem[Liu et~al.(2023)Liu, Cun, Liu, Wang, Zhang, Chen, Liu, Zeng, Chan, and Shan]{liu2023evalcrafter}
Yaofang Liu, Xiaodong Cun, Xuebo Liu, Xintao Wang, Yong Zhang, Haoxin Chen, Yang Liu, Tieyong Zeng, Raymond Chan, and Ying Shan.
\newblock Evalcrafter: Benchmarking and evaluating large video generation models, 2023.

\bibitem[Luo et~al.(2023)Luo, Chen, Zhang, Huang, Wang, Shen, Zhao, Zhou, and Tan]{luo2023videofusion}
Zhengxiong Luo, Dayou Chen, Yingya Zhang, Yan Huang, Liang Wang, Yujun Shen, Deli Zhao, Jingren Zhou, and Tieniu Tan.
\newblock Videofusion: Decomposed diffusion models for high-quality video generation.
\newblock In \emph{CVPR}, 2023.

\bibitem[Morozov et~al.(2020)Morozov, Voynov, and Babenko]{morozov2020self}
Stanislav Morozov, Andrey Voynov, and Artem Babenko.
\newblock On self-supervised image representations for gan evaluation.
\newblock In \emph{ICLR}, 2020.

\bibitem[Naeem et~al.(2020)Naeem, Oh, Uh, Choi, and Yoo]{naeem2020reliable}
Muhammad~Ferjad Naeem, Seong~Joon Oh, Youngjung Uh, Yunjey Choi, and Jaejun Yoo.
\newblock Reliable fidelity and diversity metrics for generative models.
\newblock In \emph{ICML}, pages 7176--7185. PMLR, 2020.

\bibitem[Parmar et~al.(2022)Parmar, Zhang, and Zhu]{parmar2022aliased}
Gaurav Parmar, Richard Zhang, and Jun-Yan Zhu.
\newblock On aliased resizing and surprising subtleties in gan evaluation.
\newblock In \emph{CVPR}, 2022.

\bibitem[Ramesh et~al.(2022)Ramesh, Dhariwal, Nichol, Chu, and Chen]{ramesh2022hierarchical}
Aditya Ramesh, Prafulla Dhariwal, Alex Nichol, Casey Chu, and Mark Chen.
\newblock Hierarchical text-conditional image generation with clip latents.
\newblock \emph{arXiv preprint arXiv:2204.06125}, 1\penalty0 (2):\penalty0 3, 2022.

\bibitem[Ranzato et~al.(2014)Ranzato, Szlam, Bruna, Mathieu, Collobert, and Chopra]{ranzato2014video}
MarcAurelio Ranzato, Arthur Szlam, Joan Bruna, Michael Mathieu, Ronan Collobert, and Sumit Chopra.
\newblock Video (language) modeling: a baseline for generative models of natural videos.
\newblock \emph{arXiv preprint arXiv:1412.6604}, 2014.

\bibitem[Rombach et~al.(2022)Rombach, Blattmann, Lorenz, Esser, and Ommer]{rombach2022high}
Robin Rombach, Andreas Blattmann, Dominik Lorenz, Patrick Esser, and Bj{\"o}rn Ommer.
\newblock High-resolution image synthesis with latent diffusion models.
\newblock In \emph{CVPR}, 2022.

\bibitem[R\"ossler et~al.(2019)R\"ossler, Cozzolino, Verdoliva, Riess, Thies, and Nie{\ss}ner]{roessler2019faceforensicspp}
Andreas R\"ossler, Davide Cozzolino, Luisa Verdoliva, Christian Riess, Justus Thies, and Matthias Nie{\ss}ner.
\newblock Face{F}orensics++: Learning to detect manipulated facial images.
\newblock In \emph{ICCV}, 2019.

\bibitem[Saharia et~al.(2022)Saharia, Chan, Saxena, Li, Whang, Denton, Ghasemipour, Gontijo~Lopes, Karagol~Ayan, Salimans, et~al.]{saharia2022photorealistic}
Chitwan Saharia, William Chan, Saurabh Saxena, Lala Li, Jay Whang, Emily~L Denton, Kamyar Ghasemipour, Raphael Gontijo~Lopes, Burcu Karagol~Ayan, Tim Salimans, et~al.
\newblock Photorealistic text-to-image diffusion models with deep language understanding.
\newblock In \emph{NeurIPS}, 2022.

\bibitem[Saito et~al.(2017)Saito, Matsumoto, and Saito]{saito2017temporal}
Masaki Saito, Eiichi Matsumoto, and Shunta Saito.
\newblock Temporal generative adversarial nets with singular value clipping.
\newblock In \emph{ICCV}, 2017.

\bibitem[Sajjadi et~al.(2018)Sajjadi, Bachem, Lucic, Bousquet, and Gelly]{sajjadi2018assessing}
Mehdi~SM Sajjadi, Olivier Bachem, Mario Lucic, Olivier Bousquet, and Sylvain Gelly.
\newblock Assessing generative models via precision and recall.
\newblock In \emph{NeurIPS}, 2018.

\bibitem[Salimans et~al.(2016)Salimans, Goodfellow, Zaremba, Cheung, Radford, and Chen]{salimans2016improved}
Tim Salimans, Ian Goodfellow, Wojciech Zaremba, Vicki Cheung, Alec Radford, and Xi Chen.
\newblock Improved techniques for training gans.
\newblock In \emph{NeurIPS}, 2016.

\bibitem[Schuhmann et~al.(2022)Schuhmann, Beaumont, Vencu, Gordon, Wightman, Cherti, Coombes, Katta, Mullis, Wortsman, Schramowski, Kundurthy, Crowson, Schmidt, Kaczmarczyk, and Jitsev]{schuhmann2022laionb}
Christoph Schuhmann, Romain Beaumont, Richard Vencu, Cade~W Gordon, Ross Wightman, Mehdi Cherti, Theo Coombes, Aarush Katta, Clayton Mullis, Mitchell Wortsman, Patrick Schramowski, Srivatsa~R Kundurthy, Katherine Crowson, Ludwig Schmidt, Robert Kaczmarczyk, and Jenia Jitsev.
\newblock {LAION}-5b: An open large-scale dataset for training next generation image-text models.
\newblock In \emph{Thirty-sixth Conference on Neural Information Processing Systems Datasets and Benchmarks Track}, 2022.

\bibitem[Seshadrinathan et~al.(2010)Seshadrinathan, Soundararajan, Bovik, and Cormack]{seshadrinathan2010study}
Kalpana Seshadrinathan, Rajiv Soundararajan, Alan~Conrad Bovik, and Lawrence~K Cormack.
\newblock Study of subjective and objective quality assessment of video.
\newblock \emph{IEEE transactions on Image Processing}, 19\penalty0 (6):\penalty0 1427--1441, 2010.

\bibitem[Sevilla-Lara et~al.(2021)Sevilla-Lara, Zha, Yan, Goswami, Feiszli, and Torresani]{Sevilla-Lara_2021_WACV}
Laura Sevilla-Lara, Shengxin Zha, Zhicheng Yan, Vedanuj Goswami, Matt Feiszli, and Lorenzo Torresani.
\newblock Only time can tell: Discovering temporal data for temporal modeling.
\newblock In \emph{Proceedings of the IEEE/CVF Winter Conference on Applications of Computer Vision (WACV)}, pages 535--544, 2021.

\bibitem[Shen et~al.(2023)Shen, Li, and Elhoseiny]{Shen_2023_CVPR}
Xiaoqian Shen, Xiang Li, and Mohamed Elhoseiny.
\newblock Mostgan-v: Video generation with temporal motion styles.
\newblock In \emph{CVPR}, 2023.

\bibitem[Siarohin et~al.(2019)Siarohin, Lathuilière, Tulyakov, Ricci, and Sebe]{siarohin2019first}
Aliaksandr Siarohin, Stéphane Lathuilière, Sergey Tulyakov, Elisa Ricci, and Nicu Sebe.
\newblock First order motion model for image animation.
\newblock In \emph{NeurIPS}, 2019.

\bibitem[Singer et~al.(2022)Singer, Polyak, Hayes, Yin, An, Zhang, Hu, Yang, Ashual, Gafni, et~al.]{singer2022make}
Uriel Singer, Adam Polyak, Thomas Hayes, Xi Yin, Jie An, Songyang Zhang, Qiyuan Hu, Harry Yang, Oron Ashual, Oran Gafni, et~al.
\newblock Make-a-video: Text-to-video generation without text-video data.
\newblock \emph{arXiv preprint arXiv:2209.14792}, 2022.

\bibitem[Sinno and Bovik(2018)]{sinno2018large}
Zeina Sinno and Alan~Conrad Bovik.
\newblock Large-scale study of perceptual video quality.
\newblock \emph{IEEE Transactions on Image Processing}, 28\penalty0 (2):\penalty0 612--627, 2018.

\bibitem[Skorokhodov et~al.(2022)Skorokhodov, Tulyakov, and Elhoseiny]{skorokhodov2021stylegan}
Ivan Skorokhodov, Sergey Tulyakov, and Mohamed Elhoseiny.
\newblock Stylegan-v: A continuous video generator with the price, image quality and perks of stylegan2.
\newblock In \emph{CVPR}, 2022.

\bibitem[Song et~al.(2021)Song, Meng, and Ermon]{song2021denoising}
Jiaming Song, Chenlin Meng, and Stefano Ermon.
\newblock Denoising diffusion implicit models.
\newblock In \emph{ICLR}, 2021.

\bibitem[Soomro et~al.(2012)Soomro, Zamir, and Shah]{soomro2012ucf101}
Khurram Soomro, Amir~Roshan Zamir, and Mubarak Shah.
\newblock Ucf101: A dataset of 101 human actions classes from videos in the wild.
\newblock \emph{arXiv preprint arXiv:1212.0402}, 2012.

\bibitem[Srivastava et~al.(2015)Srivastava, Mansimov, and Salakhudinov]{srivastava2015unsupervised}
Nitish Srivastava, Elman Mansimov, and Ruslan Salakhudinov.
\newblock Unsupervised learning of video representations using lstms.
\newblock In \emph{ICML}, 2015.

\bibitem[Szegedy et~al.(2016)Szegedy, Vanhoucke, Ioffe, Shlens, and Wojna]{Szegedy_2016_CVPR}
Christian Szegedy, Vincent Vanhoucke, Sergey Ioffe, Jon Shlens, and Zbigniew Wojna.
\newblock Rethinking the inception architecture for computer vision.
\newblock In \emph{CVPR}, 2016.

\bibitem[Tian et~al.(2021)Tian, Ren, Chai, Olszewski, Peng, Metaxas, and Tulyakov]{tian2021a}
Yu Tian, Jian Ren, Menglei Chai, Kyle Olszewski, Xi Peng, Dimitris~N. Metaxas, and Sergey Tulyakov.
\newblock A good image generator is what you need for high-resolution video synthesis.
\newblock In \emph{ICLR}, 2021.

\bibitem[Tong et~al.(2022)Tong, Song, Wang, and Wang]{tong2022videomae}
Zhan Tong, Yibing Song, Jue Wang, and Limin Wang.
\newblock Videomae: Masked autoencoders are data-efficient learners for self-supervised video pre-training.
\newblock \emph{NeurIPS}, 35:\penalty0 10078--10093, 2022.

\bibitem[Tulyakov et~al.(2018)Tulyakov, Liu, Yang, and Kautz]{Tulyakov_2018_CVPR}
Sergey Tulyakov, Ming-Yu Liu, Xiaodong Yang, and Jan Kautz.
\newblock Mocogan: Decomposing motion and content for video generation.
\newblock In \emph{CVPR}, 2018.

\bibitem[Unterthiner et~al.(2018)Unterthiner, Van~Steenkiste, Kurach, Marinier, Michalski, and Gelly]{unterthiner2019fvd}
Thomas Unterthiner, Sjoerd Van~Steenkiste, Karol Kurach, Raphael Marinier, Marcin Michalski, and Sylvain Gelly.
\newblock Towards accurate generative models of video: A new metric \& challenges.
\newblock \emph{arXiv preprint arXiv:1812.01717}, 2018.

\bibitem[Villegas et~al.(2017)Villegas, Yang, Hong, Lin, and Lee]{villegas2017decomposing}
Ruben Villegas, Jimei Yang, Seunghoon Hong, Xunyu Lin, and Honglak Lee.
\newblock Decomposing motion and content for natural video sequence prediction.
\newblock In \emph{ICLR}, 2017.

\bibitem[Vondrick et~al.(2016)Vondrick, Pirsiavash, and Torralba]{vondrick2016generating}
Carl Vondrick, Hamed Pirsiavash, and Antonio Torralba.
\newblock Generating videos with scene dynamics.
\newblock \emph{NeurIPS}, 2016.

\bibitem[Wang et~al.(2016)Wang, Xiong, Wang, Qiao, Lin, Tang, and Van~Gool]{wang2016temporal}
Limin Wang, Yuanjun Xiong, Zhe Wang, Yu Qiao, Dahua Lin, Xiaoou Tang, and Luc Van~Gool.
\newblock Temporal segment networks: Towards good practices for deep action recognition.
\newblock In \emph{ECCV}, 2016.

\bibitem[Wang et~al.(2023{\natexlab{a}})Wang, Huang, Zhao, Tong, He, Wang, Wang, and Qiao]{wang2023videomaev2}
Limin Wang, Bingkun Huang, Zhiyu Zhao, Zhan Tong, Yinan He, Yi Wang, Yali Wang, and Yu Qiao.
\newblock Videomae v2: Scaling video masked autoencoders with dual masking.
\newblock In \emph{CVPR}, pages 14549--14560, 2023{\natexlab{a}}.

\bibitem[Wang et~al.(2023{\natexlab{b}})Wang, Yang, Tuo, He, Zhu, Fu, and Liu]{wang2023videofactory}
Wenjing Wang, Huan Yang, Zixi Tuo, Huiguo He, Junchen Zhu, Jianlong Fu, and Jiaying Liu.
\newblock Videofactory: Swap attention in spatiotemporal diffusions for text-to-video generation.
\newblock \emph{arXiv preprint arXiv:2305.10874}, 2023{\natexlab{b}}.

\bibitem[Wang et~al.(2023{\natexlab{c}})Wang, Zhang, Zhang, Liu, Zhang, Gao, and Sang]{wang2023videolcm}
Xiang Wang, Shiwei Zhang, Han Zhang, Yu Liu, Yingya Zhang, Changxin Gao, and Nong Sang.
\newblock Videolcm: Video latent consistency model, 2023{\natexlab{c}}.

\bibitem[Wang et~al.(2020)Wang, Bilinski, Bremond, and Dantcheva]{Wang_2020_CVPR}
Yaohui Wang, Piotr Bilinski, Francois Bremond, and Antitza Dantcheva.
\newblock G3an: Disentangling appearance and motion for video generation.
\newblock In \emph{CVPR}, 2020.

\bibitem[Wang et~al.(2023{\natexlab{d}})Wang, Chen, Ma, Zhou, Huang, Wang, Yang, He, Yu, Yang, et~al.]{wang2023lavie}
Yaohui Wang, Xinyuan Chen, Xin Ma, Shangchen Zhou, Ziqi Huang, Yi Wang, Ceyuan Yang, Yinan He, Jiashuo Yu, Peiqing Yang, et~al.
\newblock Lavie: High-quality video generation with cascaded latent diffusion models.
\newblock \emph{arXiv preprint arXiv:2309.15103}, 2023{\natexlab{d}}.

\bibitem[Wang et~al.(2004)Wang, Lu, and Bovik]{wang2004video}
Zhou Wang, Ligang Lu, and Alan~C Bovik.
\newblock Video quality assessment based on structural distortion measurement.
\newblock \emph{Signal processing: Image communication}, 19\penalty0 (2):\penalty0 121--132, 2004.

\bibitem[Weissenborn et~al.(2019)Weissenborn, T{\"a}ckstr{\"o}m, and Uszkoreit]{weissenborn2019scaling}
Dirk Weissenborn, Oscar T{\"a}ckstr{\"o}m, and Jakob Uszkoreit.
\newblock Scaling autoregressive video models.
\newblock In \emph{ICLR}, 2019.

\bibitem[Wu et~al.(2021)Wu, Huang, Zhang, Li, Ji, Yang, Sapiro, and Duan]{wu2021godiva}
Chenfei Wu, Lun Huang, Qianxi Zhang, Binyang Li, Lei Ji, Fan Yang, Guillermo Sapiro, and Nan Duan.
\newblock Godiva: Generating open-domain videos from natural descriptions.
\newblock \emph{arXiv preprint arXiv:2104.14806}, 2021.

\bibitem[Xing et~al.(2023{\natexlab{a}})Xing, Xia, Zhang, Chen, Wang, Wong, and Shan]{xing2023dynamicrafter}
Jinbo Xing, Menghan Xia, Yong Zhang, Haoxin Chen, Xintao Wang, Tien-Tsin Wong, and Ying Shan.
\newblock Dynamicrafter: Animating open-domain images with video diffusion priors, 2023{\natexlab{a}}.

\bibitem[Xing et~al.(2023{\natexlab{b}})Xing, Dai, Hu, Wu, and Jiang]{xing2023simda}
Zhen Xing, Qi Dai, Han Hu, Zuxuan Wu, and Yu-Gang Jiang.
\newblock Simda: Simple diffusion adapter for efficient video generation, 2023{\natexlab{b}}.

\bibitem[Xiong et~al.(2018)Xiong, Luo, Ma, Liu, and Luo]{xiong2018learning}
Wei Xiong, Wenhan Luo, Lin Ma, Wei Liu, and Jiebo Luo.
\newblock Learning to generate time-lapse videos using multi-stage dynamic generative adversarial networks.
\newblock In \emph{CVPR}, 2018.

\bibitem[Yan et~al.(2021)Yan, Zhang, Abbeel, and Srinivas]{yan2021videogpt}
Wilson Yan, Yunzhi Zhang, Pieter Abbeel, and Aravind Srinivas.
\newblock Videogpt: Video generation using vq-vae and transformers.
\newblock \emph{arXiv preprint arXiv:2104.10157}, 2021.

\bibitem[Yan et~al.(2023)Yan, Hafner, James, and Abbeel]{yan2023temporally}
Wilson Yan, Danijar Hafner, Stephen James, and Pieter Abbeel.
\newblock Temporally consistent transformers for video generation, 2023.

\bibitem[Yu et~al.(2023{\natexlab{a}})Yu, Cheng, Sohn, Lezama, Zhang, Chang, Hauptmann, Yang, Hao, Essa, and Jiang]{Yu_2023_CVPR}
Lijun Yu, Yong Cheng, Kihyuk Sohn, Jos\'e Lezama, Han Zhang, Huiwen Chang, Alexander~G. Hauptmann, Ming-Hsuan Yang, Yuan Hao, Irfan Essa, and Lu Jiang.
\newblock Magvit: Masked generative video transformer.
\newblock In \emph{CVPR}, pages 10459--10469, 2023{\natexlab{a}}.

\bibitem[Yu et~al.(2023{\natexlab{b}})Yu, Lezama, Gundavarapu, Versari, Sohn, Minnen, Cheng, Gupta, Gu, Hauptmann, Gong, Yang, Essa, Ross, and Jiang]{yu2023language}
Lijun Yu, José Lezama, Nitesh~B. Gundavarapu, Luca Versari, Kihyuk Sohn, David Minnen, Yong Cheng, Agrim Gupta, Xiuye Gu, Alexander~G. Hauptmann, Boqing Gong, Ming-Hsuan Yang, Irfan Essa, David~A. Ross, and Lu Jiang.
\newblock Language model beats diffusion -- tokenizer is key to visual generation, 2023{\natexlab{b}}.

\bibitem[Yu et~al.(2022)Yu, Tack, Mo, Kim, Kim, Ha, and Shin]{yu2021generating}
Sihyun Yu, Jihoon Tack, Sangwoo Mo, Hyunsu Kim, Junho Kim, Jung-Woo Ha, and Jinwoo Shin.
\newblock Generating videos with dynamics-aware implicit generative adversarial networks.
\newblock In \emph{ICLR}, 2022.

\bibitem[Yu et~al.(2023{\natexlab{c}})Yu, Sohn, Kim, and Shin]{yu2023video}
Sihyun Yu, Kihyuk Sohn, Subin Kim, and Jinwoo Shin.
\newblock Video probabilistic diffusion models in projected latent space.
\newblock In \emph{CVPR}, 2023{\natexlab{c}}.

\bibitem[Zhang et~al.(2023)Zhang, Wu, Liu, Zhao, Ran, Gu, Gao, and Shou]{zhang2023show1}
David~Junhao Zhang, Jay~Zhangjie Wu, Jia-Wei Liu, Rui Zhao, Lingmin Ran, Yuchao Gu, Difei Gao, and Mike~Zheng Shou.
\newblock Show-1: Marrying pixel and latent diffusion models for text-to-video generation, 2023.

\bibitem[Zhang et~al.(2018)Zhang, Isola, Efros, Shechtman, and Wang]{Zhang_2018_CVPR}
Richard Zhang, Phillip Isola, Alexei~A. Efros, Eli Shechtman, and Oliver Wang.
\newblock The unreasonable effectiveness of deep features as a perceptual metric.
\newblock In \emph{CVPR}, 2018.

\bibitem[Zhou et~al.(2022)Zhou, Wang, Yan, Lv, Zhu, and Feng]{zhou2022magicvideo}
Daquan Zhou, Weimin Wang, Hanshu Yan, Weiwei Lv, Yizhe Zhu, and Jiashi Feng.
\newblock Magicvideo: Efficient video generation with latent diffusion models.
\newblock \emph{arXiv preprint arXiv:2211.11018}, 2022.

\bibitem[Zhou et~al.(2019)Zhou, Gordon, Krishna, Narcomey, Fei-Fei, and Bernstein]{NEURIPS2019_65699726}
Sharon Zhou, Mitchell Gordon, Ranjay Krishna, Austin Narcomey, Li~F Fei-Fei, and Michael Bernstein.
\newblock Hype: A benchmark for human eye perceptual evaluation of generative models.
\newblock In \emph{NeurIPS}, 2019.

\end{thebibliography}
}

\appendix
\clearpage
% \maketitlesupplementary

\section{Experiment Details} 

In this section, we discuss the additional details of the datasets, video generation models, and other experiment setups. 
We will release our code to compute FVD with VideoMAE-v2 backbone features and pre-computed features for commonly used video datasets.

\paragraph{Dataset.} 
We conduct our analysis on six datasets, including two widely used video understanding benchmarks Kinetics-400~\cite{carreira2017quo} and Something-Something-v2~\cite{Goyal_2017_ICCV}, three video generation benchmarks FaceForencis~\cite{roessler2019faceforensicspp}, Sky Time-lapse~\cite{xiong2018learning}, and Taichi-HD~\cite{siarohin2019first}, and the UCF-101 dataset~\cite{soomro2012ucf101} that has been used for both tasks. 

Kinetics-400~\cite{carreira2017quo} (K400) contains $267,000$ videos of $10$ seconds in $400$ action classes. Something-Something-v2~\cite{Goyal_2017_ICCV} (SSv2) consists of $220,000$ videos of $2-6$ seconds in $174$ classes of humans performing basic actions with everyday objects. UCF-101~\cite{soomro2012ucf101} has $13,320$ videos of, on average, $7$ seconds in $101$ classes of human actions.  Sky Time-lapse~\cite{xiong2018learning} (Sky) collects $2,647$ time-lapse videos of the sky in different periods and under various weather conditions. FaceForencis~\cite{roessler2019faceforensicspp} (FFS) contains $1,000$ human talking videos collected from YouTube to facilitate Deepfake detection. We follow the official instructions to process the videos to extract the face region and obtain $704$ videos. Taichi-HD~\cite{siarohin2019first} (Taichi) is a video dataset of $280$ long YouTube videos recording a person performing Taichi, which is preprocessed into $3,335$ short clips. Note that video generation models~\cite{ge2022long,yu2021generating} trained on this dataset often sample every four frames to attain larger motion in each training clip.

\paragraph{Video Generation Models.} Our generated videos are from four video generation models, DIGAN~\cite{yu2021generating}, TATS~\cite{ge2022long}, StyleGAN-v~\cite{skorokhodov2021stylegan}, and PVDM~\cite{yu2023video}. DIGAN~\cite{yu2021generating} is a GAN-based model that leverages implicit neural representations and computation-efficient discriminators. TATS~\cite{ge2022long} extends VQGAN~\cite{Esser_2021_CVPR} for long video generation by designing time-agnostic VAE and hierarchical transformer. DIGAN and TATS-base models are trained on $16$ video frames of $128\times128$ resolution. StyleGAN-v~\cite{skorokhodov2021stylegan} extends the renowned StyleGAN architecture~\cite{Karras_2019_CVPR} for video generation by employing implicit neural representations. PVDM~\cite{yu2023video} exploits a latent diffusion architecture and efficient triplane representation. StyleGAN-v and PVDM are evaluated with resolution $256\times256$ and video length $16$ and $128$. When computing FVD scores, all four methods generate $2,048$ videos. We follow StyleGAN-v~\cite{skorokhodov2021stylegan} to save the generated videos without severe JPEG compression and sample random clips from the real videos. We can reproduce the reported FVD scores, as shown in Table. 2 in the main paper.

\paragraph{Additional Implementation Deatils.} %For the synthetic experiments of quantifying
To quantify the FVD temporal sensitivity with video distortion methods, we follow the common practice~\cite{yan2021videogpt,yu2021generating,tian2021a} to sample $2,048$ clips of resolution $128\times128$ from each video dataset. 
We apply the distortion in five pre-defined corruption levels following the previous study~\cite{hendrycks2019robustness}. 
% More distorted videos of different levels can be found on our supplementary material website. 
To probe the perceptual null space in FVD, we cast the extracted features and weights to float64 to stabilize the optimization process and avoid numerical issues. 

In addition, to compute ViT encoder features of VideoMAE~\cite{wang2023videomaev2,tong2022videomae} and TimeSFormer~\cite{gberta_2021_ICML} models, we follow the convention to exploit the pre-logit features. To extract features from the pre-trained VideoMAE encoder-decoder architecture, we take the output of the penultimate layer in the encoder and average across all the patches, which uses essentially the output from the same layer as the fine-tuned VideoMAE model. To reduce memory costs when computing $\text{FVD}_{128}$ using VideoMAE-v2 models, we cast all the features to float16, as the FVD score difference between using float16 and float32 is neglectable and often less than $0.03\%$. 

All of our experiments are performed on a single NVIDIA RTX A6000 GPU except for reproducing the StyleGAN-v variants, where we follow the official receipt to train on four NVIDIA RTX A6000 GPUs.

\section{Addition Results}
\paragraph{Quantifying the temporal sensitivity of FVD.} We expand Table 1 in the main paper to include the FVD scores on the six datasets with either spatial (S) or spatiotemporal (ST) distortion using features from the I3D model, three VideoMAE-v2 variants, two TimeSformer models, and VideoMAE-v2 models in Tables~\ref{tab:spatiotemporal_more} and~\ref{tab:spatiotemporal_more2}. By inspecting the spatial FVDs computed with VideoMAE-v2 features on different datasets, we notice that they vary less than the FVD scores using the I3D features, highlighting their generalization capacity. We also explore the TimeSformer model trained on the SSv2 dataset. Compared with the one trained on the K400 dataset reported in the main paper, it is generally more sensitive to temporal quality change due to the dataset. However, it is still on par with the I3D model as both share the same supervised objective.

\paragraph{Probing the perceptual null space in FVD.} We expand Table 7 in the main paper to include FVD and FVD\textsuperscript{*} on all the models and dataset computed with the I3D model and three VideoMAE-v2 variants in Table~\ref{tab:null_space_more}. 
% We also show more visualization results of samples with the largest/smallest weights on our supplementary material website.

\begin{table*}[t]
\setlength{\tabcolsep}{3pt}
\centering
\caption{\textbf{Results of analyzing the temporal sensitivity of FVD.} We report FVDs of synthetic videos created from real videos using spatial only or spatiotemporal distortions, where the two sets produce similar frame quality as assessed by FID and only differ in temporal quality. This table includes the results of the I3D model and three VideoMAE-v2 variants.}
\label{tab:spatiotemporal_more}
\begin{tabular}{llllllll}
\toprule
Dataset & Distortion & Type & FID & FVD\textsubscript{\tiny{I3D}} & FVD\textsubscript{\tiny{VideoMAE-v2-K710}} & FVD\textsubscript{\tiny{VideoMAE-v2-SSv2}} & FVD\textsubscript{\tiny{VideoMAE-v2-PT}} \\ \midrule
\multirow{4}{*}{UCF-101} & \multirow{2}{*}{Motion Blur} & S & 133.15 & 1460.18 & 121.37 & 277.10 & 18.33  \\
 &  & ST & 133.69\scriptsize{$(+0.4\%)$} & 1705.27\scriptsize{$(+16.8\%)$} & 147.91\scriptsize{$(+21.9\%)$} & 868.31\scriptsize{$(+213.4\%)$} & 39.21\scriptsize{$(+113.8\%)$} \\
 & \multirow{2}{*}{Elastic} & S & 175.47 & 979.48 & 167.21 & 221.83 & 7.95  \\
 &  & ST & 176.46\scriptsize{$(+0.6\%)$} & 1694.95\scriptsize{$(+73.0\%)$} & 321.96\scriptsize{$(+92.5\%)$} & 1186.91\scriptsize{$(+435.0\%)$} & 58.89\scriptsize{$(+640.6\%)$} \\\hline
\multirow{4}{*}{Sky} & \multirow{2}{*}{Motion Blur} & S & 79.11 & 211.08 & 88.80 & 127.99 & 14.22  \\
 &  & ST & 79.35\scriptsize{$(+0.3\%)$} & 286.39\scriptsize{$(+35.7\%)$} & 252.01\scriptsize{$(+183.8\%)$} & 733.41\scriptsize{$(+473.0\%)$} & 35.73\scriptsize{$(+151.2\%)$} \\
 & \multirow{2}{*}{Elastic} & S & 72.32 & 149.23 & 105.04 & 142.49 & 6.97  \\
 &  & ST & 72.52\scriptsize{$(+0.3\%)$} & 333.48\scriptsize{$(+123.5\%)$} & 438.19\scriptsize{$(+317.2\%)$} & 1056.40\scriptsize{$(+641.4\%)$} & 60.60\scriptsize{$(+769.0\%)$} \\\hline
\multirow{4}{*}{FFS} & \multirow{2}{*}{Motion Blur} & S & 80.42 & 354.49 & 95.73 & 199.90 & 13.61  \\
 &  & ST & 79.57\scriptsize{$(-1.1\%)$} & 367.35\scriptsize{$(+3.6\%)$} & 178.96\scriptsize{$(+87.0\%)$} & 717.08\scriptsize{$(+258.7\%)$} & 23.75\scriptsize{$(+74.4\%)$} \\
 & \multirow{2}{*}{Elastic} & S & 161.55 & 589.07 & 192.01 & 164.82 & 11.14  \\
 &  & ST & 161.30\scriptsize{$(-0.2\%)$} & 891.50\scriptsize{$(+51.3\%)$} & 442.62\scriptsize{$(+130.5\%)$} & 969.28\scriptsize{$(+488.1\%)$} & 54.42\scriptsize{$(+388.4\%)$} \\\hline
\multirow{4}{*}{TaiChi} & \multirow{2}{*}{Motion Blur} & S & 169.76 & 1016.78 & 100.83 & 382.37 & 25.22  \\
 &  & ST & 170.10\scriptsize{$(+0.2\%)$} & 1201.35\scriptsize{$(+18.2\%)$} & 177.51\scriptsize{$(+76.0\%)$} & 1217.34\scriptsize{$(+218.4\%)$} & 47.73\scriptsize{$(+89.3\%)$} \\
 & \multirow{2}{*}{Elastic} & S & 182.99 & 688.55 & 100.93 & 161.51 & 5.81  \\
 &  & ST & 183.21\scriptsize{$(+0.1\%)$} & 1252.72\scriptsize{$(+81.9\%)$} & 372.14\scriptsize{$(+268.7\%)$} & 1467.06\scriptsize{$(+808.3\%)$} & 66.34\scriptsize{$(+1042.6\%)$} \\\hline
\multirow{4}{*}{SSv2} & \multirow{2}{*}{Motion Blur} & S & 100.65 & 594.68 & 89.31 & 144.95 & 16.96  \\
 &  & ST & 100.62\scriptsize{$(-0.0\%)$} & 678.08\scriptsize{$(+14.0\%)$} & 135.98\scriptsize{$(+52.3\%)$} & 502.09\scriptsize{$(+246.4\%)$} & 29.93\scriptsize{$(+76.5\%)$} \\
 & \multirow{2}{*}{Elastic} & S & 143.16 & 622.87 & 216.12 & 211.98 & 9.74  \\
 &  & ST & 143.91\scriptsize{$(+0.5\%)$} & 980.44\scriptsize{$(+57.4\%)$} & 351.48\scriptsize{$(+62.6\%)$} & 746.91\scriptsize{$(+252.4\%)$} & 48.07\scriptsize{$(+393.7\%)$} \\\hline
\multirow{4}{*}{K400} & \multirow{2}{*}{Motion Blur} & S & 112.22 & 996.71 & 92.11 & 257.01 & 17.67  \\
 &  & ST & 112.85\scriptsize{$(+0.6\%)$} & 1155.53\scriptsize{$(+15.9\%)$} & 126.96\scriptsize{$(+37.8\%)$} & 785.58\scriptsize{$(+205.7\%)$} & 34.34\scriptsize{$(+94.3\%)$} \\
 & \multirow{2}{*}{Elastic} & S & 146.70 & 675.53 & 151.50 & 241.15 & 8.61  \\
 &  & ST & 146.68\scriptsize{$(-0.0\%)$} & 1189.37\scriptsize{$(+76.1\%)$} & 300.02\scriptsize{$(+98.0\%)$} & 1087.20\scriptsize{$(+350.8\%)$} & 55.01\scriptsize{$(+539.0\%)$} \\\bottomrule
\end{tabular}
\end{table*}

\begin{table*}[t]
\setlength{\tabcolsep}{3pt}
\centering
\caption{\textbf{Results of analyzing the temporal sensitivity of FVD.} We report FVDs of synthetic videos created from real videos using spatial only or spatiotemporal distortions, where the two sets produce similar frame quality as assessed by FID and only differ in temporal quality. This table includes the results of the I3D model, two TimeSformer variants, and VideoMAE-v1 model.}
\label{tab:spatiotemporal_more2}
\begin{tabular}{llllllll}
\toprule
Dataset & Distortion & Type & FVD\textsubscript{I3D} & FVD\textsubscript{TimeSfomer-k400} & FVD\textsubscript{TimeSfomer-SSv2} & FVD\textsubscript{VideoMAE-v1-k400} \\ \midrule
\multirow{4}{*}{UCF-101} & \multirow{2}{*}{Motion Blur} & Spatial & 1460.18 & 265.77 & 311.85 & 26.44  \\
 &  & Spatiotemporal & 1705.27\scriptsize{$(+16.8\%)$} & 275.09\scriptsize{$(+3.5\%)$} & 336.51\scriptsize{$(+7.9\%)$} & 46.58\scriptsize{$(+76.2\%)$} \\
 & \multirow{2}{*}{Elastic} & Spatial & 979.48 & 260.65 & 261.27 & 31.82  \\
 &  & Spatiotemporal & 1694.95\scriptsize{$(+73.0\%)$} & 313.36\scriptsize{$(+20.2\%)$} & 398.29\scriptsize{$(+52.4\%)$} & 79.85\scriptsize{$(+150.9\%)$} \\\hline
\multirow{4}{*}{Sky} & \multirow{2}{*}{Motion Blur} & Spatial & 211.08 & 154.19 & 133.98 & 19.39  \\
 &  & Spatiotemporal & 286.39\scriptsize{$(+35.7\%)$} & 169.46\scriptsize{$(+9.9\%)$} & 147.69\scriptsize{$(+10.2\%)$} & 62.33\scriptsize{$(+221.4\%)$} \\
 & \multirow{2}{*}{Elastic} & Spatial & 149.23 & 123.58 & 137.43 & 23.47  \\
 &  & Spatiotemporal & 333.48\scriptsize{$(+123.5\%)$} & 186.33\scriptsize{$(+50.8\%)$} & 249.93\scriptsize{$(+81.9\%)$} & 99.02\scriptsize{$(+321.8\%)$} \\\hline
\multirow{4}{*}{FFS} & \multirow{2}{*}{Motion Blur} & Spatial & 354.49 & 240.65 & 327.82 & 21.56  \\
 &  & Spatiotemporal & 367.35\scriptsize{$(+3.6\%)$} & 241.97\scriptsize{$(+0.5\%)$} & 311.36\scriptsize{$(-5.0\%)$} & 37.32\scriptsize{$(+73.1\%)$} \\
 & \multirow{2}{*}{Elastic} & Spatial & 589.07 & 314.96 & 390.37 & 32.61  \\
 &  & Spatiotemporal & 891.50\scriptsize{$(+51.3\%)$} & 392.19\scriptsize{$(+24.5\%)$} & 472.37\scriptsize{$(+21.0\%)$} & 102.92\scriptsize{$(+215.6\%)$} \\\hline
\multirow{4}{*}{TaiChi} & \multirow{2}{*}{Motion Blur} & Spatial & 1016.78 & 342.29 & 437.08 & 26.44  \\
 &  & Spatiotemporal & 1201.35\scriptsize{$(+18.2\%)$} & 373.35\scriptsize{$(+9.1\%)$} & 499.24\scriptsize{$(+14.2\%)$} & 56.77\scriptsize{$(+114.7\%)$} \\
 & \multirow{2}{*}{Elastic} & Spatial & 688.55 & 278.37 & 276.86 & 20.88  \\
 &  & Spatiotemporal & 1252.72\scriptsize{$(+81.9\%)$} & 365.80\scriptsize{$(+31.4\%)$} & 465.57\scriptsize{$(+68.2\%)$} & 105.27\scriptsize{$(+404.2\%)$} \\\hline
\multirow{4}{*}{SSv2} & \multirow{2}{*}{Motion Blur} & Spatial & 594.68 & 166.16 & 167.52 & 21.23  \\
 &  & Spatiotemporal & 678.08\scriptsize{$(+14.0\%)$} & 169.68\scriptsize{$(+2.1\%)$} & 184.42\scriptsize{$(+10.1\%)$} & 32.65\scriptsize{$(+53.8\%)$} \\
 & \multirow{2}{*}{Elastic} & Spatial & 622.87 & 265.63 & 186.04 & 38.13  \\
 &  & Spatiotemporal & 980.44\scriptsize{$(+57.4\%)$} & 296.53\scriptsize{$(+11.6\%)$} & 245.35\scriptsize{$(+31.9\%)$} & 84.95\scriptsize{$(+122.8\%)$} \\\hline
\multirow{4}{*}{K400} & \multirow{2}{*}{Motion Blur} & Spatial & 996.71 & 203.54 & 237.63 & 18.73  \\
 &  & Spatiotemporal & 1155.53\scriptsize{$(+15.9\%)$} & 211.09\scriptsize{$(+3.7\%)$} & 254.55\scriptsize{$(+7.1\%)$} & 33.01\scriptsize{$(+76.2\%)$} \\
 & \multirow{2}{*}{Elastic} & Spatial & 675.53 & 214.95 & 206.19 & 25.40  \\
 &  & Spatiotemporal & 1189.37\scriptsize{$(+76.1\%)$} & 251.35\scriptsize{$(+16.9\%)$} & 297.65\scriptsize{$(+44.4\%)$} & 65.24\scriptsize{$(+156.9\%)$} \\ \bottomrule
\end{tabular}
\end{table*}

\begin{figure}
    \centering
    \includegraphics[width=\linewidth]{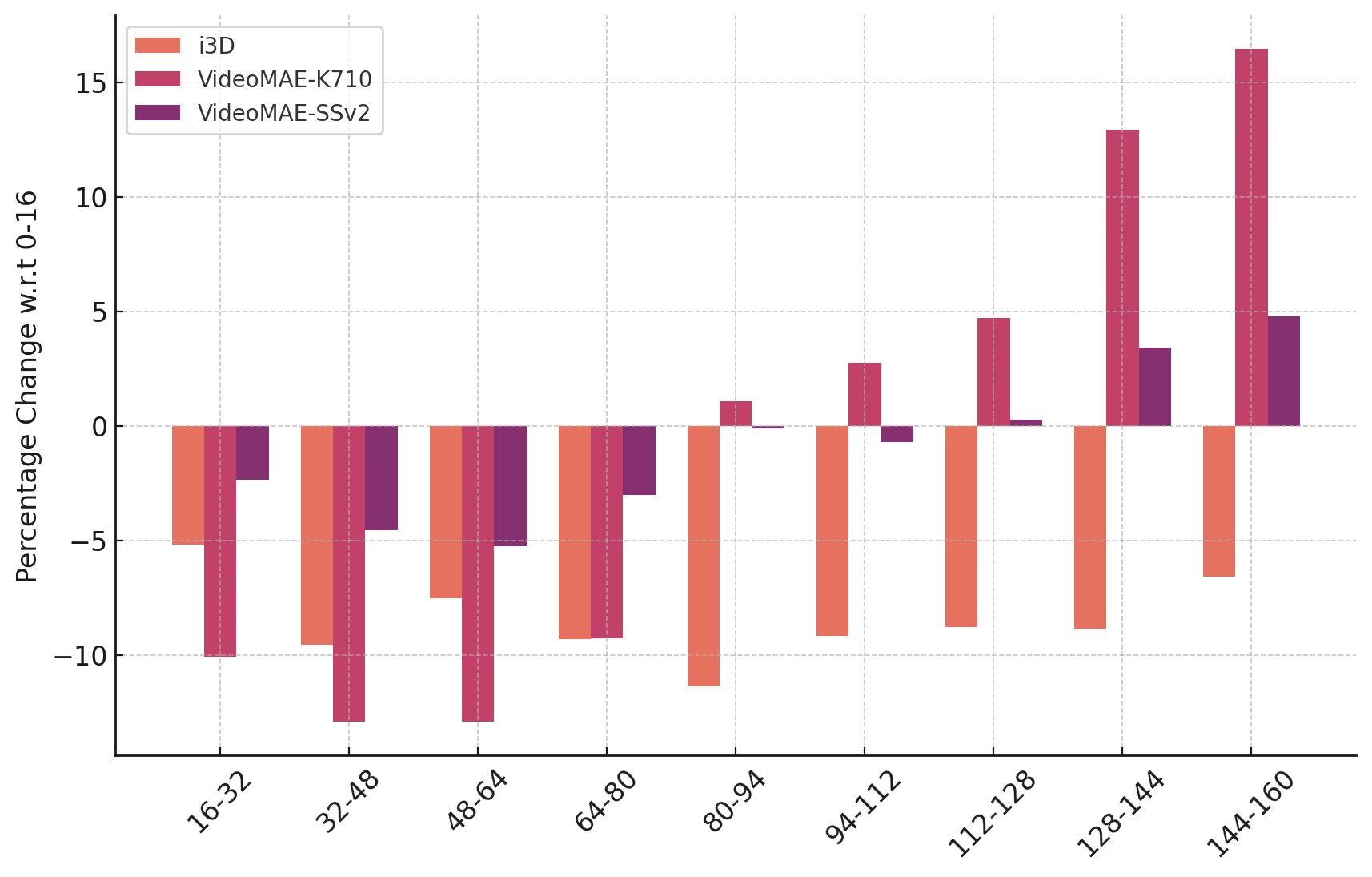}
    \caption{DIGAN~\cite{yu2021generating} trained on the Sky Time-lapse dataset generates periodic artifacts when using extrapolated time steps. We show the percentage change of FVD computed on every 16 frames compared with the first 16 frames.}
    \label{fig:frames-comp}
\end{figure}

\paragraph{Practical examples.} We expand Table 4 in the main paper by showing the FVD changes on all the consecutive 16 frames of the extrapolated generation results using DIGAN in Figure~\ref{fig:frames-comp}. We notice that with longer frames being generated, the motion artifacts become more pronounced. As a consequence, FVD scores computed with VideoMAE features properly capture the reduced temporal quality by providing a larger value. However, FVD scores computed with the I3D backbone are consistently less than from frames 0-16.
\begin{table*}[t]
    \centering
    \caption{\textbf{Results of probing the perceptual null space of FVD.} We report FVDs of normal and frozen generated videos by random sampling (FVD) and sampling to match all the fringe features (FVD\textsuperscript{*}). We color the FVD difference for better visualization: \textcolor{tablow}{$<20\%$}, \textcolor{tabmid}{$20\%-40\%$} and \textcolor{tabhigh}{$>60\%$}. The drop of FVD on the frozen generated videos indicates the volume of the null space where FVD can be reduced without generating a meaningful motion. I3D has the largest perceptual null space. }
    \label{tab:null_space_more}
    \setlength{\tabcolsep}{3pt}
    \begin{tabular}{@{}ll|cc|cc|cc|cc@{}}
    \toprule
    \multicolumn{2}{c|}{Feature Extractor} & \multicolumn{2}{c|}{I3D} & \multicolumn{2}{c|}{VideoMAE-v2-K710} & \multicolumn{2}{c|}{VideoMAE-v2-SSv2} & \multicolumn{2}{c}{VideoMAE-v2-PT} \\ 
    Model      & Dataset    & FVD     & FVD\textsuperscript{*}  & FVD\textsubscript & FVD\textsuperscript{*}  & FVD\textsubscript & FVD\textsuperscript{*} &FVD & FVD\textsuperscript{*} \\ \midrule
    \multicolumn{10}{c}{Normal Generated Videos vs. Real Videos}  \\
    DIGAN & UCF-101 & 562.36 & 220.89\scriptsize{$(\textcolor{tabhigh}{-60.7\%})$} & 358.80 & 160.13\scriptsize{$(\textcolor{tabhigh}{-55.4\%})$} & 378.19  & 260.77\scriptsize{$(\textcolor{tabmid}{-31.0\%})$} & 2.77 & 2.67\scriptsize{$(\textcolor{tablow}{-3.9\%})$} \\ 
    DIGAN & Sky & 157.13 & 54.39\scriptsize{$(\textcolor{tabhigh}{-65.4\%})$} & 86.58 & 61.93\scriptsize{$(\textcolor{tabmid}{-28.5\%})$} & 174.79  & 128.00\scriptsize{$(\textcolor{tabmid}{-26.8\%})$} & 4.72 & 3.71\scriptsize{$(\textcolor{tabmid}{-21.5\%})$} \\ 
    DIGAN & Taichi & 132.26 & 65.72\scriptsize{$(\textcolor{tabhigh}{-50.3\%})$} & 58.72 & 24.45\scriptsize{$(\textcolor{tabhigh}{-58.4\%})$} & 313.84  & 194.17\scriptsize{$(\textcolor{tabmid}{-38.1\%})$} & 4.00 & 3.66\scriptsize{$(\textcolor{tablow}{-8.5\%})$} \\ 
    TATS & UCF-101 & 329.92 & 120.58\scriptsize{$(\textcolor{tabhigh}{-63.5\%})$} & 176.98 & 72.95\scriptsize{$(\textcolor{tabhigh}{-58.8\%})$} & 388.79  & 226.39\scriptsize{$(\textcolor{tabhigh}{-41.8\%})$} & 7.92 & 7.12\scriptsize{$(\textcolor{tablow}{-10.1\%})$} \\ 
    TATS & Sky & 125.62 & 38.42\scriptsize{$(\textcolor{tabhigh}{-69.4\%})$} & 100.27 & 59.83\scriptsize{$(\textcolor{tabhigh}{-40.3\%})$} & 213.33  & 105.69\scriptsize{$(\textcolor{tabhigh}{-50.5\%})$} & 18.11 & 7.87\scriptsize{$(\textcolor{tabhigh}{-56.5\%})$} \\ 
    TATS & Taichi & 124.16 & 64.17\scriptsize{$(\textcolor{tabhigh}{-48.3\%})$} & 37.16 & 26.08\scriptsize{$(\textcolor{tabmid}{-29.8\%})$} & 274.81  & 126.53\scriptsize{$(\textcolor{tabhigh}{-54.0\%})$} & 5.88 & 5.34\scriptsize{$(\textcolor{tablow}{-9.2\%})$} \\  
    StyleGAN-V & Sky & 56.63 & 31.73\scriptsize{$(\textcolor{tabhigh}{-44.0\%})$} & 180.97 & 55.54\scriptsize{$(\textcolor{tabhigh}{-69.3\%})$} & 219.85  & 148.11\scriptsize{$(\textcolor{tabmid}{-32.6\%})$} & 10.04 & 8.78\scriptsize{$(\textcolor{tablow}{-12.5\%})$} \\ 
    StyleGAN-V & FFS & 56.22 & 25.87\scriptsize{$(\textcolor{tabhigh}{-54.0\%})$} & 77.28 & 61.02\scriptsize{$(\textcolor{tabmid}{-21.0\%})$} & 194.68  & 135.30\scriptsize{$(\textcolor{tabmid}{-30.5\%})$} & 1.08 & 1.04\scriptsize{$(\textcolor{tablow}{-3.7\%})$} \\ 
    PVDM & UCF-101 & 348.81 & 113.99\scriptsize{$(\textcolor{tabhigh}{-67.3\%})$} & 116.01 & 90.40\scriptsize{$(\textcolor{tabmid}{-22.1\%})$} & 369.14  & 172.35\scriptsize{$(\textcolor{tabhigh}{-53.3\%})$} & 4.51 & 3.69\scriptsize{$(\textcolor{tablow}{-18.2\%})$} \\ 
    PVDM & Sky & 59.95 & 22.94\scriptsize{$(\textcolor{tabhigh}{-61.7\%})$} & 141.48 & 75.12\scriptsize{$(\textcolor{tabhigh}{-46.9\%})$} & 142.50  & 57.04\scriptsize{$(\textcolor{tabhigh}{-60.0\%})$} & 3.63 & 2.33\scriptsize{$(\textcolor{tabmid}{-35.7\%})$}\\\hline
    \multicolumn{10}{c}{Frozen Generated Videos vs. Real Videos}  \\
    DIGAN & UCF-101 & 1303.13 & 715.96\scriptsize{$(\textcolor{tabhigh}{-45.1\%})$} & 357.61 & 175.13\scriptsize{$(\textcolor{tabhigh}{-51.0\%})$} & 951.59  & 859.57\scriptsize{$(\textcolor{tablow}{-9.7\%})$} & 12.61 & 12.23\scriptsize{$(\textcolor{tablow}{-3.1\%})$}\\ 
    DIGAN & Sky & 230.64 & 115.55\scriptsize{$(\textcolor{tabhigh}{-49.9\%})$} & 175.47 & 142.86\scriptsize{$(\textcolor{tablow}{-18.6\%})$} & 408.17  & 362.84\scriptsize{$(\textcolor{tablow}{-11.1\%})$} & 13.23 & 12.16\scriptsize{$(\textcolor{tablow}{-8.1\%})$}\\ 
    DIGAN & Taichi & 461.79 & 276.88\scriptsize{$(\textcolor{tabhigh}{-40.0\%})$} & 132.96 & 52.00\scriptsize{$(\textcolor{tabhigh}{-60.9\%})$} & 578.61  & 523.20\scriptsize{$(\textcolor{tablow}{-9.6\%})$} & 4.40 & 4.18\scriptsize{$(\textcolor{tablow}{-4.9\%})$}\\ 
    TATS & UCF-101 & 1157.69 & 616.25\scriptsize{$(\textcolor{tabhigh}{-46.8\%})$} & 247.80 & 107.41\scriptsize{$(\textcolor{tabhigh}{-56.7\%})$} & 908.95  & 805.88\scriptsize{$(\textcolor{tablow}{-11.3\%})$} & 14.66 & 13.68\scriptsize{$(\textcolor{tablow}{-6.7\%})$}\\ 
    TATS & Sky & 279.75 & 126.32\scriptsize{$(\textcolor{tabhigh}{-54.8\%})$} & 172.00 & 140.37\scriptsize{$(\textcolor{tablow}{-18.4\%})$} & 375.74  & 353.15\scriptsize{$(\textcolor{tablow}{-6.0\%})$} & 21.28 & 15.76\scriptsize{$(\textcolor{tabmid}{-25.9\%})$}\\ 
    TATS & Taichi & 475.99 & 312.19\scriptsize{$(\textcolor{tabmid}{-34.4\%})$} & 164.58 & 64.69\scriptsize{$(\textcolor{tabhigh}{-60.7\%})$} & 587.31  & 530.86\scriptsize{$(\textcolor{tablow}{-9.6\%})$} & 4.63 & 4.42\scriptsize{$(\textcolor{tablow}{-4.4\%})$}\\  
    StyleGAN-V & Sky & 206.56 & 104.27\scriptsize{$(\textcolor{tabhigh}{-49.5\%})$} & 224.80 & 91.71\scriptsize{$(\textcolor{tabhigh}{-59.2\%})$} & 503.22  & 456.24\scriptsize{$(\textcolor{tablow}{-9.3\%})$} & 23.17 & 21.60\scriptsize{$(\textcolor{tablow}{-6.8\%})$}\\ 
    StyleGAN-V & FFS & 353.79 & 242.04\scriptsize{$(\textcolor{tabmid}{-31.6\%})$} & 171.38 & 147.76\scriptsize{$(\textcolor{tablow}{-13.8\%})$} & 547.24  & 520.98\scriptsize{$(\textcolor{tablow}{-4.8\%})$} & 14.08 & 14.22\scriptsize{$(\textcolor{tablow}{+0.9\%})$}\\ 
    PVDM & UCF-101 & 1135.61 & 605.09\scriptsize{$(\textcolor{tabhigh}{-46.7\%})$} & 250.52 & 211.34\scriptsize{$(\textcolor{tablow}{-15.6\%})$} & 1032.90  & 898.48\scriptsize{$(\textcolor{tablow}{-13.0\%})$} & 12.95 & 12.34\scriptsize{$(\textcolor{tablow}{-4.7\%})$}\\ 
    PVDM & Sky & 182.77 & 94.87\scriptsize{$(\textcolor{tabhigh}{-48.1\%})$} & 198.50 & 140.77\scriptsize{$(\textcolor{tabmid}{-29.1\%})$} & 429.06  & 395.79\scriptsize{$(\textcolor{tablow}{-7.8\%})$} & 11.54 & 11.03\scriptsize{$(\textcolor{tablow}{-4.4\%})$} \\
    \bottomrule
    \end{tabular}
\end{table*}


\end{document}